\pdfoutput=1

\documentclass[11pt]{article}

\usepackage[]{acl}

\usepackage{times}
\usepackage{latexsym}

\usepackage[T1]{fontenc}

\usepackage[utf8]{inputenc}

\usepackage{microtype}

\usepackage{inconsolata}

\usepackage{xcolor}

\usepackage{graphicx}
\usepackage{tabularx}
\usepackage{amsmath}
\usepackage{amssymb}
\usepackage{booktabs}
\usepackage{multirow}
\usepackage{caption}
\usepackage{subcaption}
\usepackage{algorithm}
\usepackage{algpseudocode}
\usepackage{amsfonts}
\usepackage{verbatim}

\title{\emph{What the Weight?!} \\ A Unified Framework for Zero-Shot Knowledge Composition}

\author{Carolin Holtermann\textsuperscript{1}, Markus Frohmann\textsuperscript{2}, Navid Rekabsaz\textsuperscript{2}, Anne Lauscher\textsuperscript{1} \\
\textsuperscript{1}Data Science Group, University of Hamburg, Germany \\
  \textsuperscript{2}Johannes Kepler University Linz, Austria \\
  \texttt{\{carolin.holtermann,anne.lauscher\}@uni-hamburg.de} \\
  \texttt{\{markus.frohmann,navid.rekabsaz\}@jku.at} \\}

\begin{document}
 \maketitle
\begin{abstract}

The knowledge encapsulated in a model is the core factor determining its final performance on downstream tasks. 
Much research in NLP has focused on efficient methods for storing and adapting different types of knowledge, e.g., in dedicated modularized structures, and on how to effectively combine these, e.g., by learning additional parameters. However, given the many possible options, a thorough understanding of the mechanisms involved in these compositions is missing, and hence it remains unclear which strategies to utilize. 
To address this research gap, we propose a novel framework for zero-shot module composition, which encompasses existing and some novel variations for selecting, weighting, and combining parameter modules under a single unified notion. Focusing on the scenario of domain knowledge and adapter layers, our framework provides a systematic unification of concepts, allowing us to conduct the first comprehensive benchmarking study of various zero-shot knowledge composition strategies. In particular, we test two module combination methods and five selection and weighting strategies for their effectiveness and efficiency in an extensive experimental setup. Our results highlight the efficacy of ensembling but also hint at the power of simple though often-ignored weighting methods. Further in-depth analyses allow us to understand the role of weighting vs. top-k selection, and show that, to a certain extent, the performance of adapter composition can even be predicted. %

\end{abstract}

\section{Introduction}
Pre-trained language models (PLMs), e.g., the GPT-family~\citep[\emph{inter alia}]{GPT2,NEURIPS2020_1457c0d6}, determine the current state-of-the-art in Natural Language Processing (NLP), which has often been attributed to the rich knowledge they encapsulate in their parameters~\citep[e.g.,][]{tenney-etal-2019-bert}. 
Previous research has heavily focused on utilizing the PLMs' knowledge in various scenarios, particularly in a zero-shot setting, e.g., to transfer the knowledge of different source domains to a specific target domain~\citep[e.g.,][\emph{inter alia}]{emelin-etal-2022-injecting,hung-etal-2022-ds}.%

\begin{figure}[t]
    \centering
    \includegraphics[width=\linewidth]{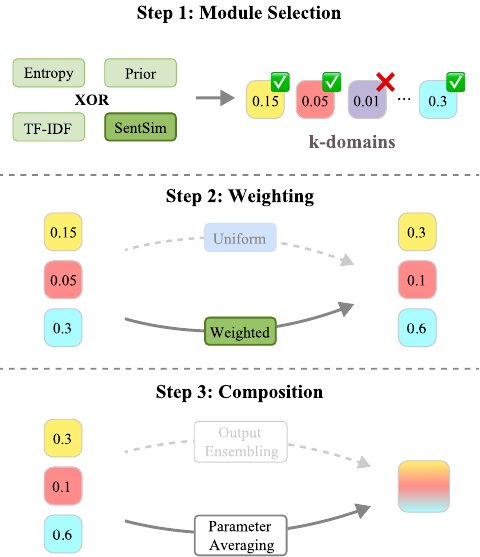}
    \caption[Module Composition Strategies]{Our unified framework for on-demand module composition consisting of three steps: selection, weighting, and final combination. We show the example of zero-shot domain adaptation with adapter layers.} %
    \label{fig:strategies}
\end{figure}
Besides the numerous practical advantages of knowledge modularization -- such as parameter-efficiency~\citep{ponti-etal-2023-combining}, avoiding catastrophic forgetting~\citep{ansell-etal-2021-mad-g}, and reducing negative interference~\citep{Sun_Shao_Li_Liu_Yan_Qiu_Huang_2020} -- researchers have shown the benefits of re-using and re-combining already existing modules~\citep{pfeiffer-etal-2021-adapterfusion}.

Based on this idea, a particularly attractive scenario is the \emph{on-demand selection and combination of knowledge modules at inference time}. %
To do so, there exist a plethora of potential strategies: modules can be selected by computing sentence similarities and domain clusters~\citep{adapterSoup}, domain priors \citep{li2022branchtrainmerge}, and model entropy \citep{wang-etal-2022-adamix}. Then, they can be combined with a weight space averaging, following the idea of a ``model soup'' \citep{wortsman2022model}, or output vector ensembling \citep{li2022branchtrainmerge}.

However, despite the existence of a variety of knowledge composition methods, there is (a)~no comprehensive overview and evaluation of those methods, and (b)~no unified view on knowledge composition that could facilitate this process. The composition methods introduced for various objectives have not been tested in a comparable setup (e.g., \citet{li2022branchtrainmerge}, do not focus on zero-shot domain adaptation, in contrast to \citet{adapterSoup}), and various factors (e.g., the number of modules to select, and whether to additionally weight each module in the composition) have not been systematically taken into account. We shed light on these, focusing on the specific case of zero-shot domain adaptation with adapter layers. Given a series of adapters originating from domain-specific training, we address the problem of how to choose and combine adapters to improve the performance on unseen evaluation domains. 
\paragraph{Contributions.} Our contributions are three-fold: \textbf{(1)}~we present a unified framework for zero-shot knowledge composition (see Figure~\ref{fig:strategies}), which provides an interoperable notion on knowledge composition variations proposed for diverse scenarios in the literature. Our framework allows us \textbf{(2)}~to conduct a large evaluation of knowledge composition strategies for zero-shot domain adaptation to date. Concretely, we test two combination methods (averaging and ensembling), and five selection and weighting strategies (uniform, and based on model entropy, domain prior, semantic sentence similarity, and TF--IDF (which has been previously ignored) across three models ({\small\texttt{gpt2-base}}, {\small\texttt{gpt2-large}}, {\small\texttt{deberta-base}}) using 21 training and 10 evaluation domains. 
\textbf{(3)}~We advance our understanding of knowledge composition by proposing and studying a meta-regression method applied to the framework, aiming to predict the optimal combinatorial setting. 

Our experiments show that w.r.t. combination strategies, output vector ensembling is often superior to parameter averaging, supporting findings from recent work \citep{li2022branchtrainmerge}. Importantly, we observe that corpus-based weighting and selection strategies (\textsc{tf--idf} and \textsc{sentence similarity}) often outperform more complex model-based approaches, while also being more efficient. Our study on meta-regression shows that zero-shot domain adaptation performance is partially predictable, particularly for specific adapter combinations. We hope that our work will advance efficient and effective NLP. For full reproducibility, we release all code publicly under \url{https://github.com/UhhDS/WhatTheWeight}.

\section{A Unified Composition Framework}
In this section, we present our unified framework for knowledge module composition. We base our explanation on the scenario of domain adaptation using adapters as the underlying module. Our framework is, however, generic and can be applied to various composition scenarios. 

The problem of composing knowledge boils down to the following: let  $\theta_i$ be the parameters of $n$ adapters trained via language modeling on $n$ domains $D_1,...,D_n$ while the original model parameters $\phi$ are kept frozen. Given an unseen evaluation domain $D_{n+1}$, the task is to effectively adapt to $D_{n+1}$ via an optimal domain composition. As illustrated in Figure \ref{fig:strategies}, our approach to such a composition relies on three steps: (\textbf{1})~identify $k$ suitable adapters; (\textbf{2})~apply a weighting to the selected adapters; (\textbf{3})~perform the final combination. In the following, we describe the scoring and the combination strategies, implemented in our framework and used for conducting the experiments. %

\subsection{Scoring Strategy}
We examine five scoring strategies. These strategies are utilized for selecting the top-$k$ most suitable adapters (\textbf{1}), and/or to compute the weights $\omega_{i}$ per domain (\textbf{2}) which will later be used in the combination. %
Concretely, our framework consists of uniform,  %
two corpus-based, and two model-based scoring approaches, explained in the following.

\paragraph{Uniform.} In this simplest method (\textsc{uniform}), the scores follow a uniform distribution with values of $ \omega_i = 1/k$. 
This strategy can not be used for selecting the top-$k$, but %
it can be paired with other strategies that provide the top-$k$ best domain adapters, by further weighting these uniformly.

\paragraph{Semantic Sentence Similarity.} This is a corpus-based scoring strategy (\textsc{SentSim}). In line with \citet{adapterSoup}, we compute Sentence-BERT \citep{reimers-gurevych-2019-sentence} embeddings for 100 randomly selected sequences of the development set of each of the training domains $D_1,...,D_n$, and of the unseen evaluation domain $D_{n+1}$. Next, we compute the averaged cosine similarity for each $D_1,...,D_n$ across the 100 training embeddings with each of the 100 embeddings from $D_{n+1}$. We obtain the final \textsc{SentSim} scores through normalization, dividing each cosine similarity by the sum of all similarities. The resulting scores are in $[0,1]$, such that $\sum_{i=1}^k \omega_i = 1$. %

\paragraph{TF--IDF.} In contrast to previous work, we also examine Term Frequency--Inverse Document Frequency (\textsc{tf--idf}), as another simple corpus-based scoring strategy. Here, we are motivated by the fact that domain differences also manifest in different lexical choices. As before, we extract 100 sequences of the development sets of each of the training domains and of the novel evaluation domain. We then compute TF--IDF vectors for each subset and compute the scores as the normalized average cosine similarity (see above). We provide the exact TF--IDF formulation in the Appendix \ref{sec:tfidf_equation}.

\paragraph{Domain Prior.}
Following \citet{demix} and \citet{li2022branchtrainmerge}, here, we consider score estimation as a Bayesian problem (\textsc{prior}): we introduce a domain variable $D$ alongside each sequence $x$ of the evaluation set and define $p(x|D=j)$ as the conditional probability of the last token in the sequence, given the preceding tokens, calculated by applying a softmax over the model output vector. Applying Bayes' rule, we estimate the domain posterior $p(D=j | x)$ (the probability of a sequence belonging to the domain $j$) as follows:

\begin{small}
\begin{equation}
\begin{aligned}
p(D=j | x) &= \frac{p(x|D=j) \cdot p(D=j)}{p(x)} \\
 &= \frac{p(x|D=j) \cdot p(D=j)}{\sum_{j'=1}^k p(x|D=j') \cdot p(D=j')}\,.
\end{aligned}
\end{equation}
\end{small}

\noindent To estimate the domain prior ${P(D=j)}$, we compute the exponential moving average (EMA) of the posterior probabilities at the end of each sequence block. We use $N = 100$ sequences of the dev sets with a sequence length of 1024 and an EMA decay of $\lambda = 0.3$, which has been found to result in stable posterior probabilities \citep{li2022branchtrainmerge}. 

\begin{small}
\begin{equation}%
p(D=j) = \sum_{i=1}^N \lambda^i \cdot p(D=j|x^{(i)})\,,
\end{equation}
\end{small}

\noindent with individual input sequences $x_i$. We then fix the obtained domain priors and use those as scores at inference time. We apply averaging normalization, causing the scores of $k$ adapters to sum up to 1.

\paragraph{Entropy.} This method leverages model uncertainty as a scoring strategy (\textsc{entropy}). Our method has conceptual similarities to the one of \citet{emea}, %
while in contrast instead of running multiple gradient descent iterations, we opt for a more efficient strategy and %
measure the uncertainty for each adapter on the development sets $X$ with a single pass. %
Similar to \citet{uncertainty}, we define model uncertainty as the entropy of the predicted probability distribution:

\small{
\begin{equation}%
    H(X) = -\sum_{x\in X} p(x) \cdot \log p(x)\,,
\end{equation}}
\normalsize

\noindent with mini-batches $x$, and $p(x)$ being the mean probability of the next token given the preceding tokens for all sequences in the batch. For each adapter, we then compute the uncertainty of the model on the evaluation set (that is, the data corresponding to the unseen domain). The resulting uncertainties are then normalized to obtain certainty scores with values in the range of $[0,1]$. This way, the domain adapter achieving the lowest uncertainty on the evaluation set gets the highest weight assigned.

\subsection{Combination Method}
Given the weight vector $\omega$ we obtained from steps (\textbf{1}) and (\textbf{2}), we rely on two combination methods to combine the knowledge modules (\textbf{3}).

\paragraph{Parameter Averaging.} %
We follow \citet{adapterSoup} and use ``model souping'' \citep{wortsman2022model}, namely weight space averaging, as our first combination strategy. %
To ensure consistency, we also treat the parameters of the PLM heads of auto-encoding models as parts of $\theta_i$ -- the parameters specific to a particular domain $D_i$, as these appear to have a major impact on the down-stream task. Here, we thus average over both the adapter layers and the weight space of the head's parameters. %
Expanding on the original proposal by \citet{adapterSoup}, we also allow for the weighting of the adapters. %
In particular, we consider $f(x, \phi, \theta_i)$ as a single model with its original parameters $\phi$, and the domain-specific adapter and head parameters $\theta_i$ operating on the provided textual input $x$. The new model using the parameter averaging method is hence formulated as: 

\small{
\begin{equation}%
    f(x, \phi, \sum_{i=1}^k \omega_i * \theta_i)\,,
\end{equation}}
\normalsize

\noindent with $\omega_i$ as the weight for the domain-specific parameters $\theta_i$, and $k$ the number of selected adapters. %

\paragraph{Ensembling.} In this method, we ensemble the outputs of $k$ selected models $f(x, \phi, \theta_i)$, each defined with the corresponding domain-specific parameters. This strategy is similar to the one proposed in \citet{li2022branchtrainmerge}.%

\small{
\begin{equation}%
    \sum_{i=1}^k \omega_i * f(x, \phi, \theta_i)\,.
\end{equation}}
\normalsize

\noindent Compared to averaging, this strategy requires a separate pass through each model of the ensemble. %

\section{Benchmarking Composition Strategies} \label{sec:benchmark}
We use our framework to benchmark module composition strategies for zero-shot domain adaptation.
\subsection{Overall Experimental Setup} \label{sec:experimental_setup}
\paragraph{Data.} We follow \citet{adapterSoup} and resort to defining domains by provenance, i.e., the source of a document. Although the notion of a domain is fuzzy \citep{plank,domain-adaptation}, the document sources provide an intuitive segmentation of the corpora while also being common practice in NLP research. %
We use the same 21 training domains, which correspond to collections of text from 21 websites, and 10 evaluation domains as in \citep{adapterSoup}. 30 of these constitute domains from the 100 most high-resource internet domains from the %
C4 dataset \citep{C4,Cleaned-C4}. We also add the publicly available \texttt{yelp.com} dataset.\footnote{https://www.yelp.com/dataset} We show all datasets along with their train-eval split sizes in Table \ref{tab:datasets}.

\setlength{\tabcolsep}{12.5pt}
\begin{table}[t!]
    \centering
    \small
    \begin{tabular}{llr}
        \toprule
          \textbf{Split} & \textbf{Datasets} & \textbf{%
          \# Tokens} \\
        \midrule
        \multirow{21}{*}{Train} &  dailymail.co.uk  & 23M (3M)\\
        & wired.com  & 18M (2M)\\
        & express.co.uk & 13M (2M) \\
        & npr.org  & 24M (3M)\\
        & librarything.com  & 2M (300K)\\
        &  instructables.com  & 24M (3M)\\
        & entrepreneur.com  & 15M (2M)\\
        & link.springer.com  & 23M (3M)\\
        & insiderpages.com  & 6M (700K)\\
        & ign.com  & 9M (1M)\\
        & eventbrite.com  & 6M (800K)\\
        & forums.macrumors.com  & 19M (2M)\\
        & androidheadlines.com  & 14M (2M)\\
        & glassdoor.com  & 2M (200K)\\
        & pcworld.com  & 13M (2M)\\
        &  csmonitor.com  & 22M (3M)\\
        &  lonelyplanet.com  & 4M (500K)\\
        &  booking.com  & 30M (4M)\\
        &   journals.plos.org  & 6M (1M)\\
        &   frontiersin.org  & 31M (4M)\\
        &   medium  & 21M (3M)\\
        \hline
        \multirow{10}{*}{Eval} & reuters.com  & 16M (2M)\\
        & techcrunch.com  & 12M (2M)\\
        & fastcompany.com  & 13M (2M)\\
        &  nme.com  & 3M (300K)\\
        &  fool.com  & 34M (4M)\\
        &  inquisitr.com  & 13M (2M)\\
        & mashable.com & 12M (2M)\\
        & tripadvisor.com  & 5M (1M) \\
        & ncbi.nlm.nih.gov & 21M (3M)\\
        & yelp.com & 15M (2M)\\
        \bottomrule
    \end{tabular}
    \normalsize
    \caption{Datasets used in our study. We show the 21 training and 10 evaluation domains with their sizes measured in number of tokens (training (eval)).}
    \label{tab:datasets}
\end{table}

\paragraph{Models.} We evaluate one auto-encoding and two auto-regressive models. %
To be able to compare our results to \citet{adapterSoup}, we use GPT-2 \citep{GPT2} in the \textit{base} configuration ({\small\texttt{gpt2-base}}). Additionally, we evaluate the \textit{large} configuration ({\small\texttt{gpt2-large}}) and further  train domain adapters for the DeBERTa model \citep{deberta} in the \textit{base} configuration ({\small\texttt{deberta-base}}). We obtain all models from the Huggingface Transformers library \citep{wolf-etal-2020-transformers}. 

\paragraph{Adapter Training and Optimization.}
We train each domain adapter separately via language modeling (masked language modeling or causal language modeling, depending on the model) on a single NVIDIA A6000 GPU with 48 GB RAM. For each adapter, we use a random seed of $5$ during training. %
We train for 20 epochs using the Adam optimizer~\citep{AdamW} (weight decay = $0.01$, $\beta_1 = 0.9$, $\beta_2 = 0.999$, $\epsilon = 1 \cdot 10^{-6}$, learning rate=$1\cdot 10^{-4}$). %
For {\small\texttt{deberta-base}} and {\small\texttt{gpt2-base}}, we use %
an effective batch size of 80, while for the bigger model, {\small\texttt{gpt2-large}}, we set the effective batch size to 20. %
To make the results of {\small\texttt{gpt2-base}} comparable to the results of \citet{adapterSoup}, we adopt the adapter architecture proposed by \citet{bapna-firat-2019-simple}, that is, we insert an adapter layer after the transformer feed-forward layer. We set the reduction factor to $12$, resulting in a bottleneck size of 64 for {\small\texttt{gpt2-base}} and {\small\texttt{deberta-base}}, and 107 for {\small\texttt{gpt2-large}}.

\paragraph{Evaluation.} For each evaluation domain, we measure the models' perplexities obtained after adapter composition. 
All evaluations are conducted over 4 different random seeds (${5, 10, 42, 88}$) and averaged to achieve stable results.

\begin{figure}[t!]
     \centering
     \begin{subfigure}{0.5\textwidth}
         \centering
         \includegraphics[width=0.8\linewidth]{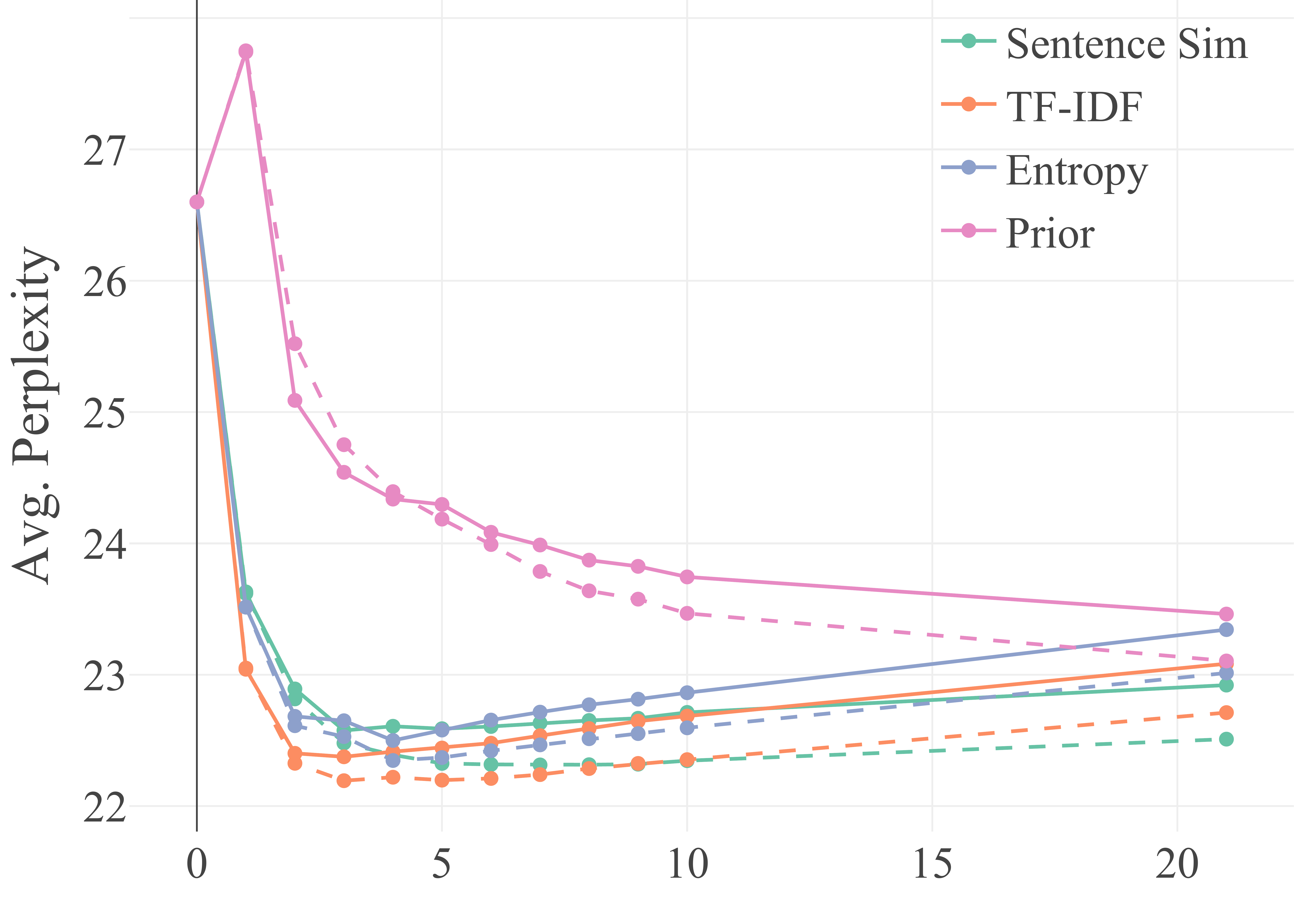}
         \caption{{\small\texttt{gpt2-base}}}
     \end{subfigure}
     \begin{subfigure}{0.5\textwidth}
         \centering
         \includegraphics[width=0.8\linewidth]{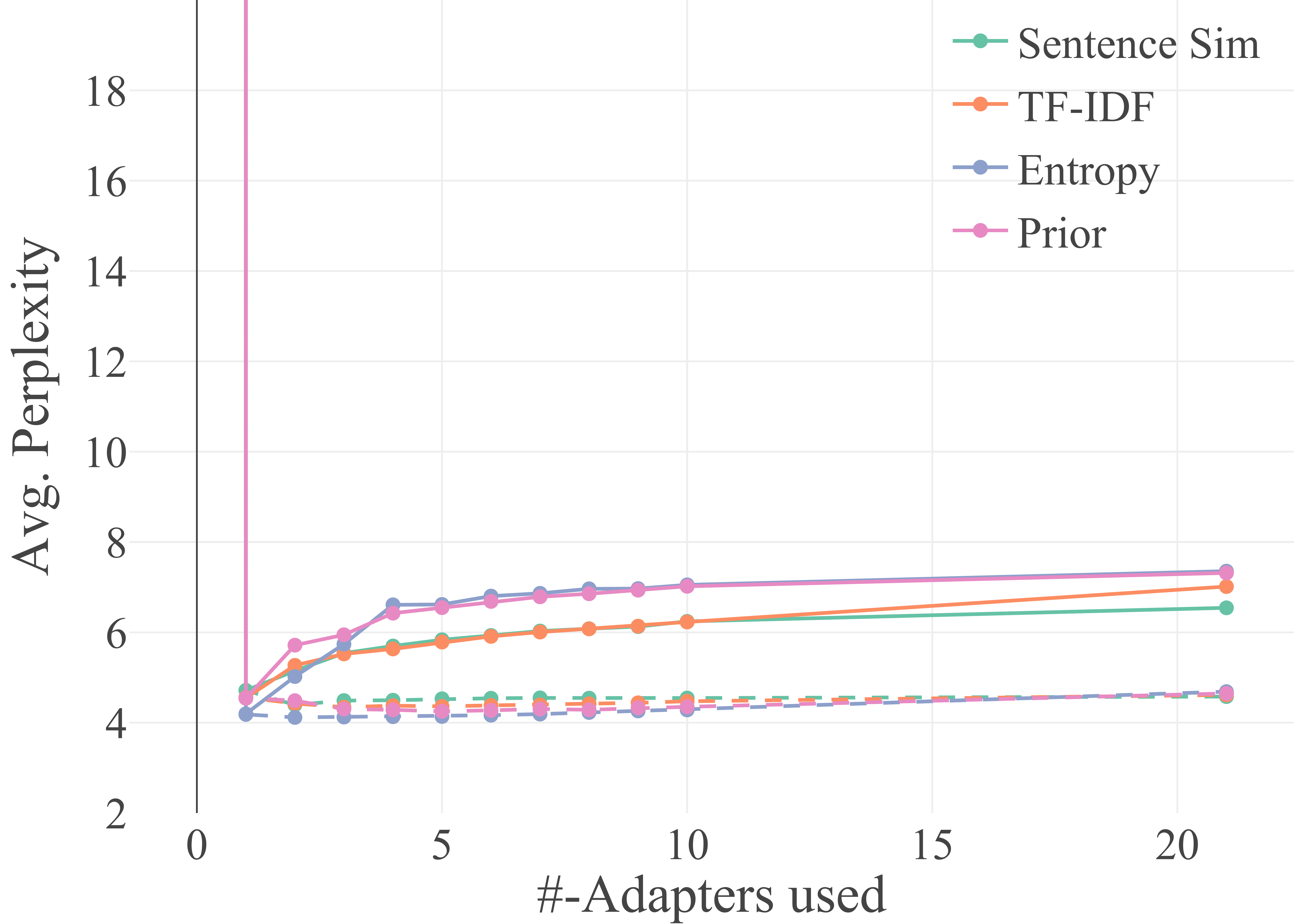}
         \caption{{\small\texttt{deberta-base}}}
     \end{subfigure}
        \caption{Comparison between Parameter Averaging (solid lines) and Ensembling (dashed lines) over different numbers of top-$k$ adapters. We show the mean perplexity results for (a) {\small\texttt{gpt2-base}}, and (b) {\small\texttt{deberta-base}} for each of our scoring strategies (\textsc{SentSim}, \textsc{tf--idf}, \textsc{entropy}, \textsc{prior}) averaged across four runs.}
        \label{fig:AVG_ENS_Comp}
\end{figure}
\subsection{Results}
\paragraph{Combination Strategies.} We compare the two combination strategies, parameter averaging, and ensembling, coupled with all four scoring strategies, applied for adapter selection and adapter weighting. The perplexities for {\small\texttt{gpt2-base}} and {\small\texttt{deberta-base}} are depicted in Figure~\ref{fig:AVG_ENS_Comp}. We show results for {\small\texttt{gpt2-large}} in the Appendix \ref{sec:app_avg_ens}. Note that for $k=0$ and $k=1$ (no adapter or a single adapter), the combination strategies are equivalent, as we do not need to merge any adapters. Interestingly, {\small\texttt{deberta-base}} hugely profits from adding a single adapter (improvement of up to -183662.70 in perplexity). Adding a second adapter does, on average, when averaging modules, no longer lead to an improvement. This warrants further investigation on when exactly the knowledge contained in an adapter helps (cf. \S\ref{sec:meta}). From $k=2$ on, ensembling leads to better domain adaptation across most model types and scoring strategies, indicated by lower model perplexities. These findings hold when choosing two adapters only ($k=2$) and also when increasing $k$, up to $k=21$ (all adapters chosen) and are significant at $\alpha = 0.05$ using the Wilcoxon Signed Rank test. With larger $k$ the difference between the combination strategies even increases (from -0.08 for $k=2$ to -0.41 for $k=21$ and \textsc{tf--idf}). The only exception is \texttt{prior} for {\small\texttt{gpt2-base}}, where averaging reaches better performance for smaller $k$. Overall, we can confirm the recent findings of \citet{li2022branchtrainmerge}: ensembling typically leads to better performance than module averaging. Beyond plain performance aspects, we also note that ensembling shows wider applicability than parameter averaging, concretely, when diverse adapter architectures are involved. However, we also conclude that adding more adapters can also harm the performance.

\paragraph{Scoring Strategies.}

\setlength{\tabcolsep}{3pt}
\begin{table*}[t]
    \centering
    \small
    \begin{tabular}{llrrrrrrrrrr}
        \toprule
        & \multicolumn{11}{c}{\textbf{Results on the 10 Evaluation Domains (AVG/ENS)}} \\
          &\textbf{Method} & \textbf{reuters} & \textbf{techcru} & \textbf{fastco} & \textbf{nme} & \textbf{fool} & \textbf{inquisitr} & \textbf{mashable} & \textbf{tripadv} & \textbf{ncbi} & \textbf{yelp}  \\
        \midrule
        \parbox[t]{2mm}{\multirow{2}{*}{\rotatebox[origin=c]{90}{$\spadesuit$}}}& &  21.5 & 27.7 & 27.9 & 28.2 & 23.8 & 22.4 & 27.1 & 40.4 & 20.7 & 36.2 \\
        & \textsc{SentSim} &  17.6 & 22.0 & 21.3 & 20.7 & 22.2 & 18.4 & 22.4 & 36.2 & 17.6 & 35.2 \\
        \midrule
        \parbox[t]{2mm}{\multirow{6}{*}{\rotatebox[origin=c]{90}{{\small\texttt{gpt2-base}}}}} & & 20.2 & 27.4 &  27.1 & 28.4 & 22.9 & 21.9 & 25.7 & 38.4 & 19.7 & 34.4 \\
         &\textsc{uniform} & 16.9/16.4 &23.2/22.6 &22.8/21.9 &22.8/21.9 &21.3/21.3 &18.3/17.3 &22.2/21.9 &34.6/33.8 &18.2/18.0 &33.3/34.4 \\
         & \textsc{SentSim} &  \textbf{16.5/16.1} & \textbf{22.8/22.3} & \textbf{22.5}/21.7 & 22.3/\textbf{21.5} & \textbf{21.2/21.2} & \textbf{18.0/17.6} & \textbf{21.9/21.6} & \textbf{33.7/32.4} & \textbf{17.4/17.2} & \textbf{32.9/33.7} \\
         &\textsc{tf--idf} & 16.5/16.1 &22.8/22.3 &22.5/\textbf{21.7} & \textbf{22.2/21.5} &21.3/21.2 &18.0/17.6 &22.1/21.7 &34.4/33.4 &17.8/17.5 &33.2/34.1 \\
         &\textsc{entropy} & 16.8/16.4 &23.2/22.6 &22.8/21.9 &22.8/21.9 &21.3/21.3 &18.3/17.8 &22.3/21.9 &34.6/33.8 &18.2/18.0 &33.3/34.4 \\
         &\textsc{prior} & 17.1/16.6 &23.4/22.8 &23.1/22.2 &23.1/22.3 &21.4/21.4 &18.4/18.0 &22.4/22.1 &34.4/33.6 &18.2/18.1 &33.2/34.2 \\
        \midrule
        \parbox[t]{2mm}{\multirow{6}{*}{\rotatebox[origin=c]{90}{{\small\texttt{gpt2-large}}}}} & & 12.2 & 17.5 &  17.1 & 16.6 & 15.4 & 14.0 & 16.7 & 26.4 & 12.6 & \textbf{23.0} \\
        & \textsc{uniform} & 11.2/10.6 &16.0/15.3 & 15.5/14.8 &14.6/13.7 &14.9/14.4 &12.7/12.1 &15.3/14.6 &24.2/23.2 &11.9/11.7 &24.0/23.5 \\
        & \textsc{SentSim} &  11.1/\textbf{10.5} & \textbf{15.7/15.0} & 15.4/\textbf{14.7} & \textbf{14.3/13.5} & \textbf{14.9/14.4} & \textbf{12.5/12.0} & \textbf{15.1/14.4} & \textbf{23.3/22.2} & \textbf{11.4/11.1} & 23.3/23.6 \\
        & \textsc{tf--idf} & \textbf{11.1/10.5} &15.8/15.1 & \textbf{15.4}/14.7 &14.3/13.5 &14.9/14.4 &12.5/12.0 &15.2/14.5 &24.0/22.9 &11.7/11.3 &23.8/23.9 \\
        & \textsc{entropy} &11.2/10.8 &16.0/15.5 &15.5/15.0 &14.6/14.0 &14.9/14.6 &12.7/12.3 &15.3/14.6 &24.2/23.2 &11.9/11.7 &24.0/24.2 \\
        & \textsc{prior} &11.2/10.7 &16.1/15.4 &15.6/14.9 &14.7/13.9 &14.9/14.5 &12.7/12.2 &15.3/14.7 &24.1/23.0 &11.9/11.7 &23.9/24.1 \\
        \midrule
        \parbox[t]{2mm}{\multirow{6}{*}{\rotatebox[origin=c]{90}{{\small\texttt{deberta-base}}}}} &  & 116975.5 & 123763.4 &  122145.2 & 117231.9 & 125070.4 & 118561.9 & 118559.0 & 123046.6 & 110694.9 & 125107.5 \\
        & \textsc{uniform} & 6.7/4.1 &7.1/4.5 &6.4/\textbf{4.1} &7.1/4.6 &7.1/\textbf{4.4} &5.8/3.7 &6.8/4.2 &9.8/\textbf{6.3} &8.8/5.8 &8.4/5.5  \\
        & \textsc{SentSim} &  \textbf{5.9/3.9}	& \textbf{6.3/4.4}	& \textbf{5.9}/\textbf{4.1}	&\textbf{6.2}/\textbf{4.5}	& \textbf{6.4}/\textbf{4.4} &\textbf{5.1}/\textbf{3.5}	& \textbf{6.1}/4.2 &\textbf{8.7}/\textbf{6.3}	& \textbf{7.0}/\textbf{4.6} &\textbf{7.9}/5.8\\
        & \textsc{tf--idf} &  6.2/4.0 & 6.6/\textbf{4.4} & 6.1/\textbf{4.1} & 6.6/\textbf{4.5} & 6.8/\textbf{4.4} &5.4/3.6 & 6.5/4.2 & 9.4/\textbf{6.3} & 8.4/5.2 &8.2/5.5 \\
        & \textsc{entropy}&  6.6/4.0 & 7.1/\textbf{4.4} &6.4/\textbf{4.1} &7.0/4.6 & 7.0/\textbf{4.4} &5.7/3.6 &6.8/4.2 &9.8/\textbf{6.3} &8.7/6.3 &8.4/5.5 \\
        & \textsc{prior}& 6.6/4.0 &6.9/\textbf{4.4} & 6.4/\textbf{4.1} &7.0/\textbf{4.5} &7.0/\textbf{4.4} &5.6/3.6 &6.7/\textbf{4.2} &9.8/\textbf{6.3} &8.7/5.6 &8.4/\textbf{5.4} \\

        \bottomrule
    \end{tabular}
    \caption{Perplexity results obtained when using all trained adapters for prediction on an evaluation domain. We compare the different scoring (\textsc{uniform}, \textsc{SentSim}, \textsc{TF--IDF}, \textsc{Entropy}, and \textsc{Prior}) and combination strategies (parameter averaging (AVG) and output ensembling (ENS)) averaged over 4 different initializations. The perplexities marked with $\spadesuit$ represent the results of \citet{adapterSoup} obtained with {\small\texttt{gpt2-base}}.}
    \label{tab:overall_results}
\end{table*}

\begin{figure}[t]
     \centering
     \begin{subfigure}{0.5\textwidth}
         \centering
         \includegraphics[width=.8\linewidth]{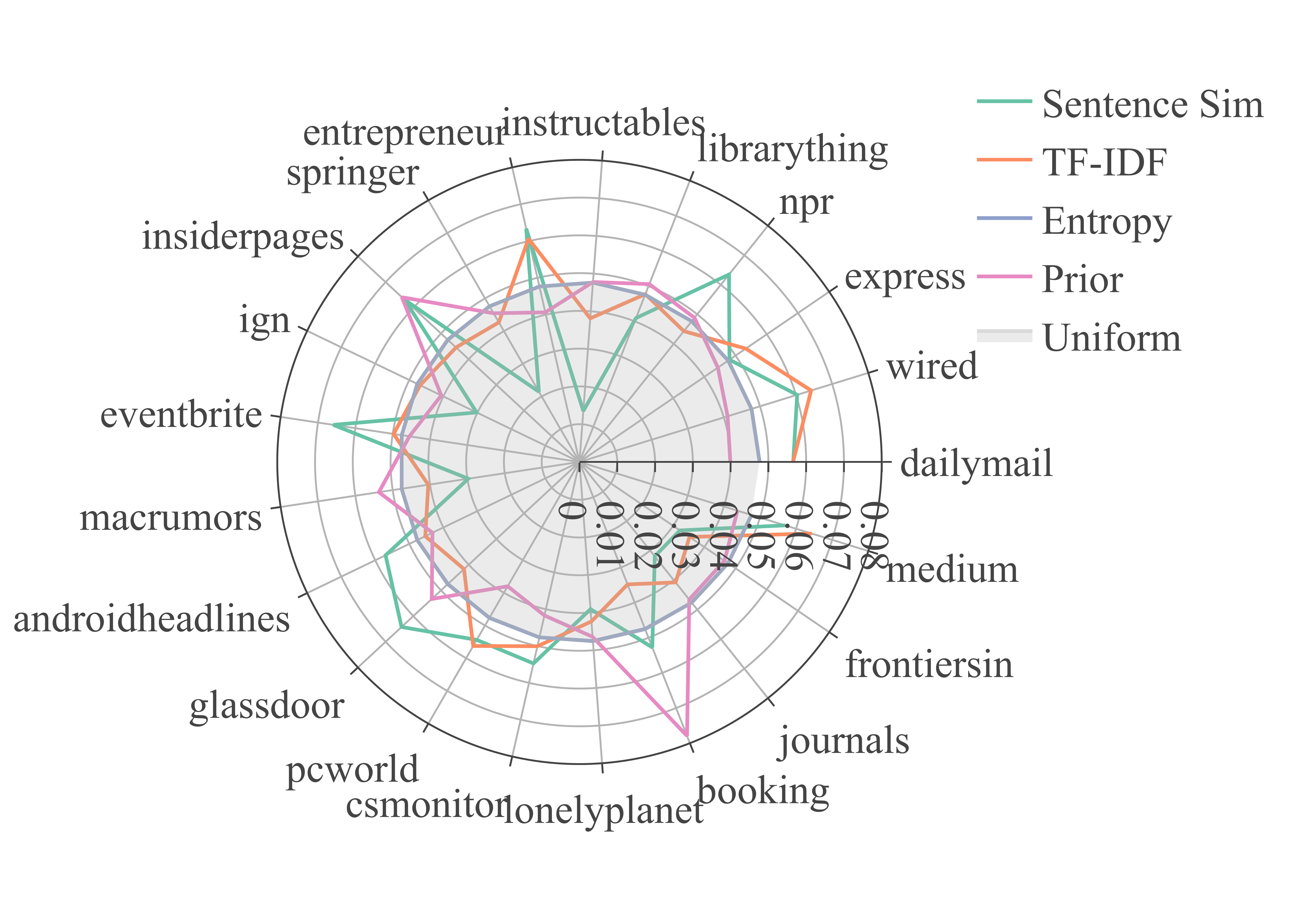}
         \caption{{\small\texttt{gpt2-base}}}
     \end{subfigure}
     \begin{subfigure}{0.5\textwidth}
         \centering
         \includegraphics[width=.7\linewidth]{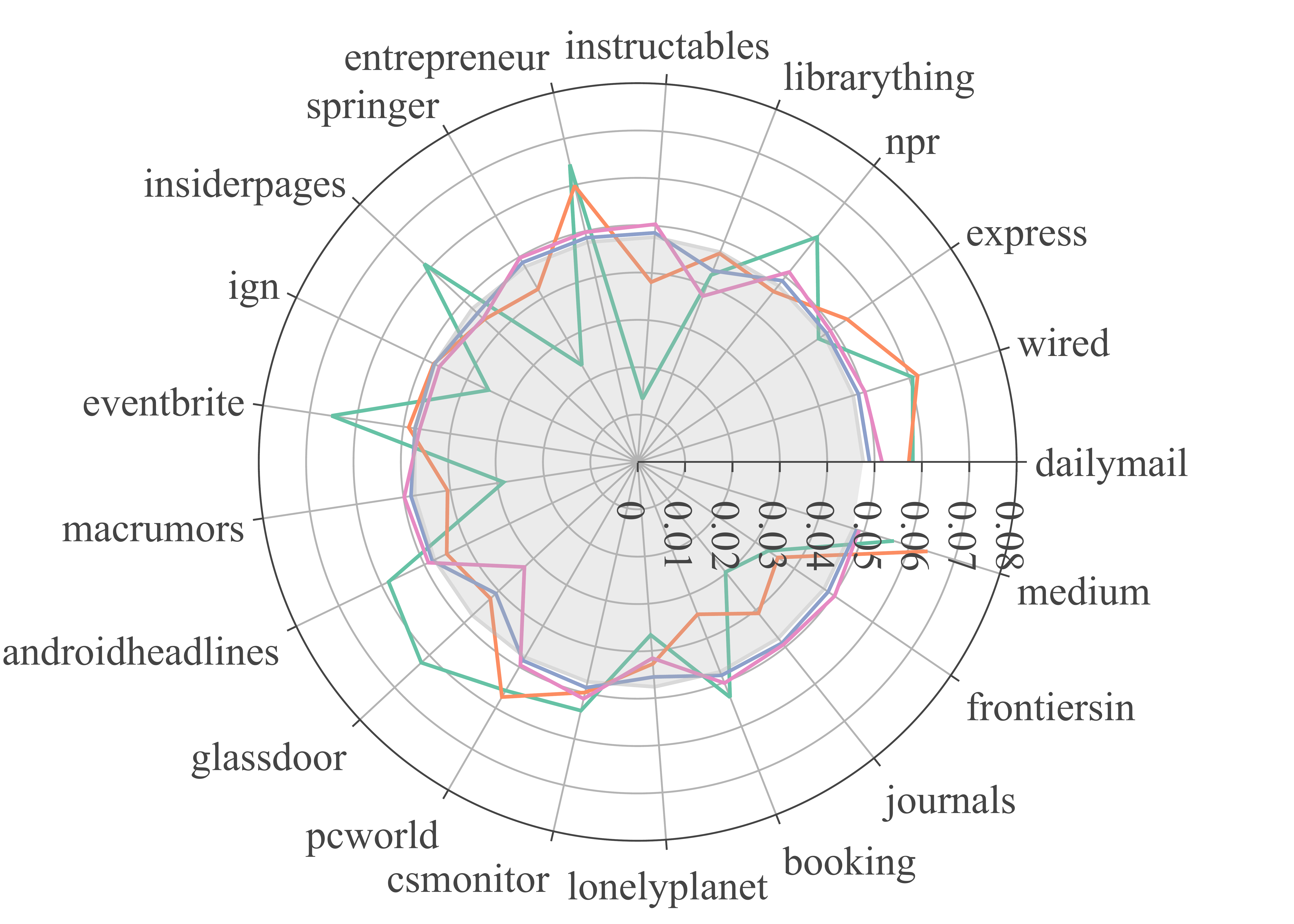}
         \caption{{\small\texttt{deberta-base}}}
     \end{subfigure}
        \caption{Adapter weights for all training domains and scoring strategies when using all trained adapters. The light grey shade indicates the uniform weighting. }
        \label{fig:WEIGHTS}
\end{figure}
\begin{figure}[t!]
     \centering
     \begin{subfigure}{0.235\textwidth}
         \centering
         \includegraphics[width=1.\linewidth]{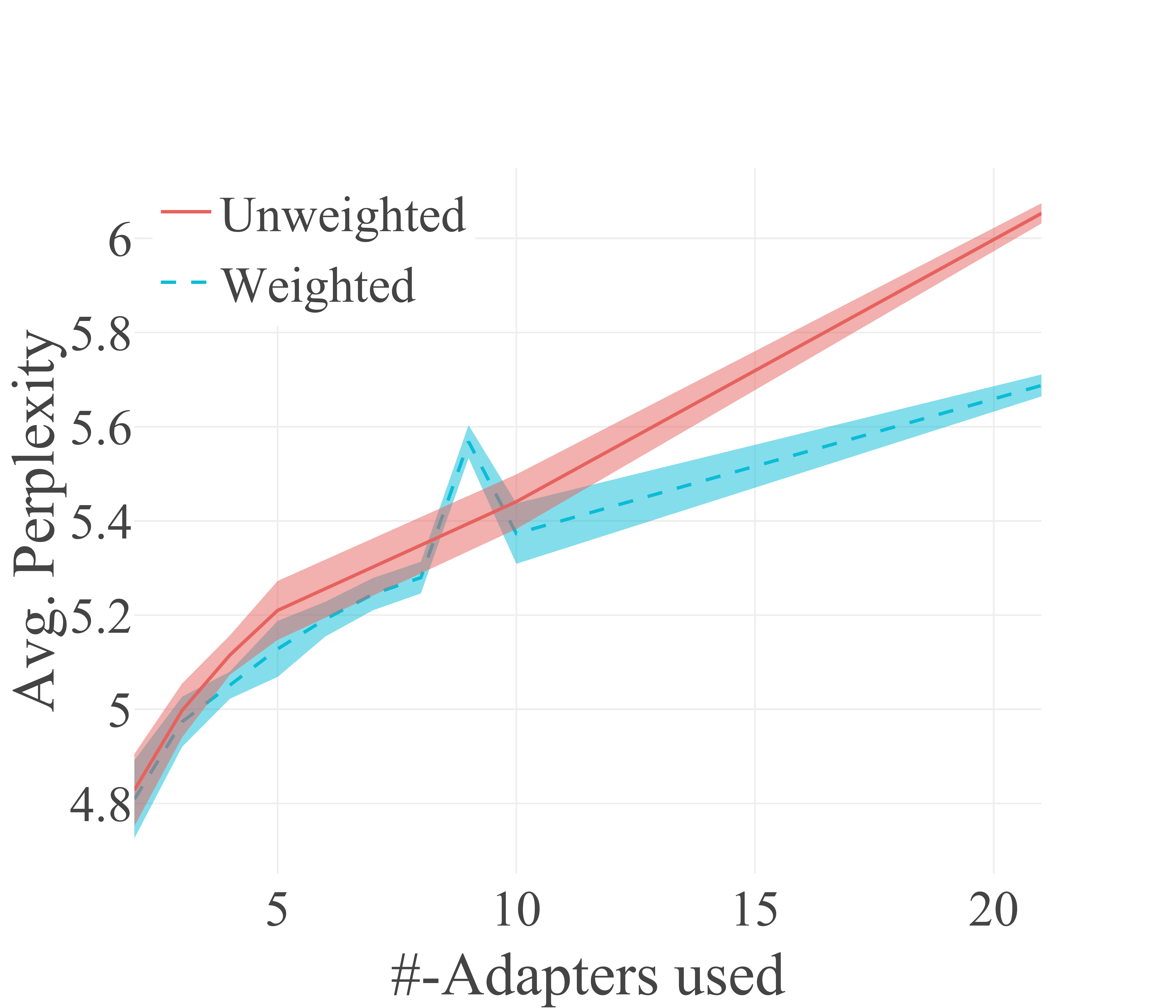}
         \caption{{\small\texttt{deberta-base}}}
     \end{subfigure}
     \begin{subfigure}{0.235\textwidth}
         \centering
         \includegraphics[width=1.\linewidth]{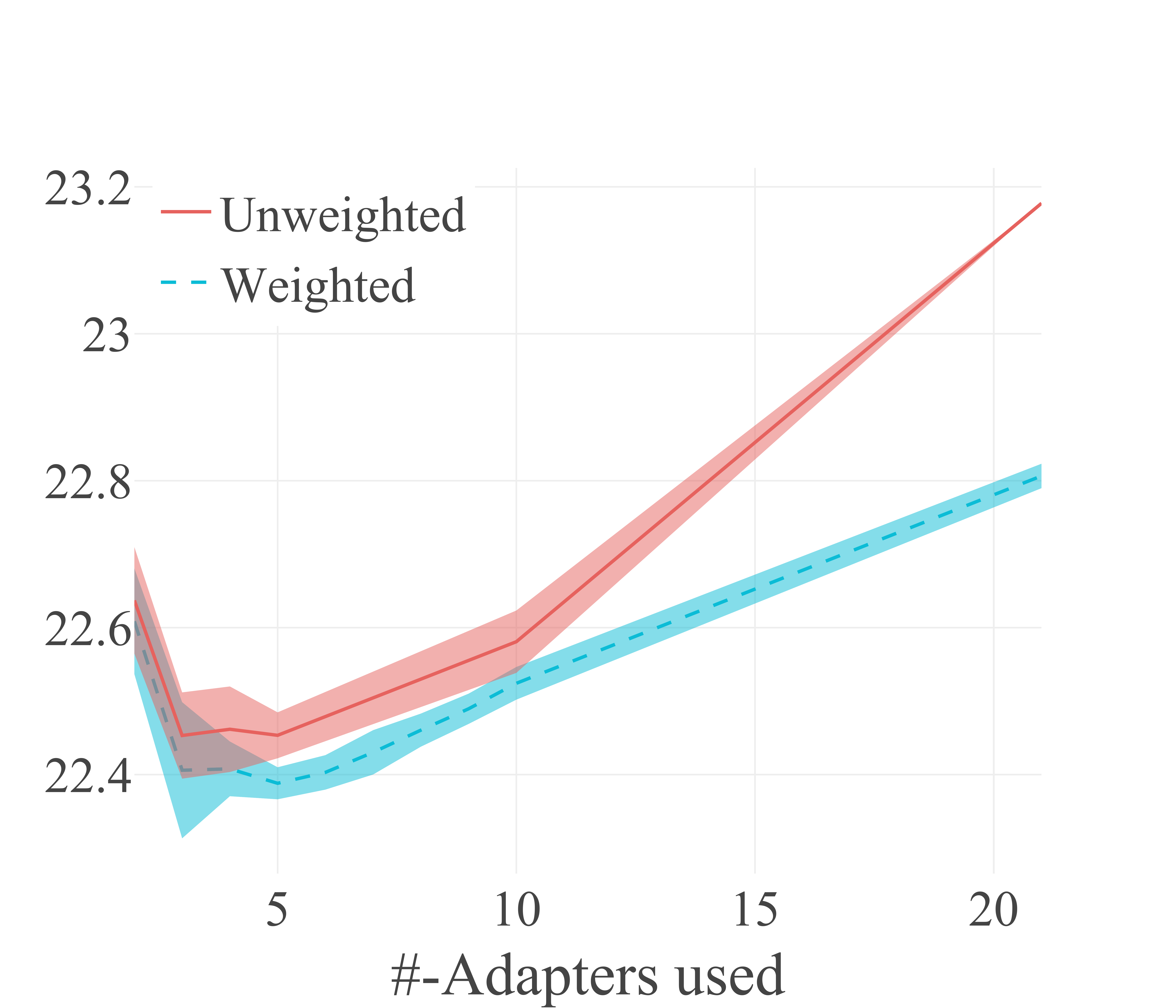}
         \caption{{\small\texttt{gpt2-base}}}
     \end{subfigure}
        \caption{Comparison between weighting adapters based on their similarity (blue) and assigning them uniform weights (red). We show the mean perplexity results for (a) {\small\texttt{deberta-base}}, and (b) {\small\texttt{gpt2-base}} and when using corpus-based scoring strategies (\textsc{tf--idf}, \textsc{SentSim}) averaged over four runs and both combination strategies.}
        \label{fig:weighted-unweighted}
\end{figure}

We evaluate the effectiveness of the scoring strategies for weighting all 21 training adapters (see Table~\ref{tab:overall_results}).
Surprisingly, we observe that simpler (and previously ignored) approaches to determine the weighting, e.g., \textsc{SentSim} and \textsc{tf--idf}, often lead to better results compared to more sophisticated approaches. However, for smaller numbers of adapters, the picture can vary (see again Figure~\ref{fig:AVG_ENS_Comp}). To shed more light on this phenomenon, we show the weights obtained through the different scoring strategies in Figure~\ref{fig:WEIGHTS}: the model-based scoring strategies produce weight distributions closer to the uniform distribution than the two corpus-based ones, where domain differences are more pronounced. We conclude that model-based ones are thus, while providing good results in adapter selection (i.e., when a fixed and smaller $k$ is chosen), less suitable for fine-grained weighting of a larger set of adapters. 
We are also interested in whether the more advanced scoring strategies should be used as weighting mechanisms or whether uniform weighting leads to superior results. To this end, we compute the perplexities on all evaluation datasets in two variants: (i)~when using the different scoring strategies (e.g., \textsc{tf--idf})  for selection and weighting, and (ii)~when only using them for selection and then uniformly weighting the selected adapters. As already indicated by the weight differences depicted in Figure~\ref{fig:WEIGHTS}, we do not expect big differences for model-based strategies (e.g., \textsc{entropy}). However, for the corpus-based strategies, weighting has a small but visible effect (up to 0.3711 for $k=21$). We show the average scores obtained across all evaluation datasets and across these strategies (\textsc{tf--idf} and \textsc{SentSim}) in Figure \ref{fig:weighted-unweighted}: for higher $k$, weighting generally has a positive impact. It can thus be an alternative to fixing $k$ -- removing this additional hyperparameter -- for the corpus-based scoring strategies. %
Yet, selecting a good number of adapters still stands out as a more crucial factor for optimal performance.

\paragraph{Efficiency.} A particular motivation for modularization is the re-usability of the individual modules -- leading to a reduction of the environmental impact~\citep{Strubell_Ganesh_McCallum_2020, hershcovich-etal-2022-towards}. Here, we discuss the efficiency of the combination strategies we test within our framework. As pointed out by \citet{li2022branchtrainmerge}, ensembling is intrinsically more expensive at inference time than averaging -- the amount of parameters is linearly increasing with the number of modules added. We now measure the expected CO$_2$ equivalents in our concrete experimental setup. This complements our understanding of the fine-grained differences among the individual scoring strategies. Following \citet{hershcovich-etal-2022-towards}, we compute the CO$_2$ equivalents in gram ($\text{gCO$_2$eq}$) as follows:

\begin{small}
\begin{equation}
\begin{aligned}
\text{gCO$_2$eq} = \\
\text{ComputationTime (hours)} \times \\ 
\text{Power(kW)} \times \\ 
\text{EnergyMix (gCO$_2$eq/kWh)} 
\end{aligned}
\end{equation}
\end{small}

\begin{figure}[t!]
    \centering
    \includegraphics[width=.85\linewidth]{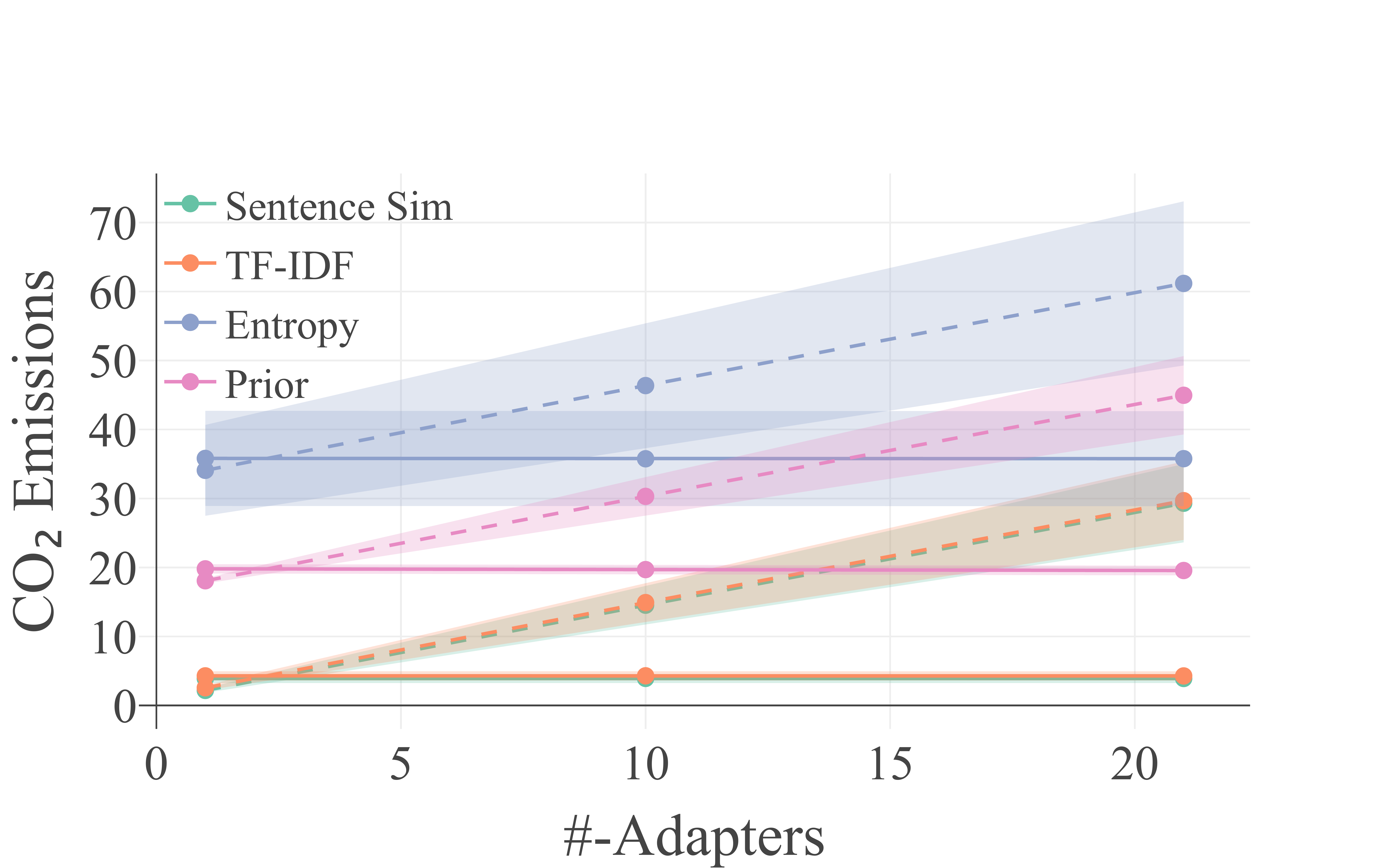}
    \caption{The different scoring and combination strategies with regards to their efficiency. We show the results for {\small\texttt{gpt2-base}} for Parameter Averaging (solid lines) and Ensembling (dashed lines) paired with each of our four scoring strategies and averaged across four runs.}
    \label{fig:carbon_emissions}
\end{figure}

\noindent We estimate these by measuring the computation time needed for each selection paired with each selection strategy. All experiments are carried out on a single NVIDIA A6000 GPU (TDP 300W) except for the score calculations with \textsc{tf--idf} and \textsc{SentSim}. These were run on a single AMD EPYC 7313 CPU (TDP 155W). We employ a private server infrastructure located in Germany with a carbon intensity of 470g.\footnote{Estimate from \url{https://app.electricitymaps.com/zone/DE}} We compute the mean carbon emission across 4 initialization seeds and display the results in Figure \ref{fig:carbon_emissions}.

As expected, we measure a linear increase for ensembling, while averaging does not result in increased CO$_2$ equivalents. Unsurprisingly, the model-based strategies are more expensive than the corpus-based ones. Here, \textsc{entropy}-based selection results in the highest amount of estimated carbon emissions (up to 61.17 gCO$_2$ vs. 3.91 for \textsc{tf--idf} and ensembling). %

\section{Meta-Regression} \label{sec:meta}
In \S\ref{sec:benchmark}, we have shown that adding more adapters (i.e., increasing $k$) often does not lead to performance gains, and that the effectiveness of the scoring strategies varies across models and evaluation domains. Motivated by these results, here, we analyze to what extent we are able to predict the expected performance for particular compositions.

\subsection{Experimental Setup} 
\paragraph{Dataset and Evaluation.} We run a meta-regression on our results obtained for each base model in \S\ref{sec:benchmark}. We pre-process the data as follows: to account for variations in the scores, we average over the results obtained from the four random seeds for each evaluation domain. We account for the base differences in perplexity among the evaluation domains by computing the delta between the original model performance on this dataset and the perplexity obtained by using the composition, normalized by the original perplexity. We use 10-fold cross-validation and report the results in terms of Pearson and Spearman Correlation.

\paragraph{Features.} Each instance is represented by  five feature groups: \textit{Adapter} -- the weights assigned to particular training adapters ($0$ if not chosen); \textit{Number of  Adapters} -- the number of adapters involved in the composition; \textit{Combination Strategy} -- one-hot encoding of average or ensembling;  \textit{Scoring Strategy} -- one-hot encodings of the scoring strategies (e.g., \textsc{tf--idf}); and \textit{Evaluation Dataset} -- one-hot encodings of the target domain.  %

\paragraph{Models and Baselines.}  We experiment with Linear and Ridge regression. For Ridge, we perform hyperparameter tuning ($\alpha$), leading to $\alpha=0$ for {\small\texttt{gpt2-base}}, $\alpha=0.17$ for {\small\texttt{deberta-base}} and $\alpha=0.06$ for {\small\texttt{gpt2-large}}. We compare the results with a baseline predicting the mean relative difference per evaluation dataset. We hypothesize this to be a strong baseline, as the effectiveness of an adapter combination is highly dependent on the target domain. 

\paragraph{Results.} %
Both models surpass the baseline (see Table \ref{tab:metaregression}), which, as expected, already reaches high scores. The highest scores are achieved with Ridge regression on the {\small\texttt{gpt2-base}} results ($0.9641$ Spearman). The results on {\small\texttt{deberta-base}} are the lowest, indicating the model type to be a relevant factor. Overall, we conclude that, dependent on the PLM, we are able to predict the effectiveness of domain adaptation with various compositions if metadata from previous studies can be leveraged. This finding holds promise for reducing the time and resources required for extensive experimental evaluation, for instance, when an organization seeks to expand an existing approach to a novel application domain (e.g., a startup focusing on the intersection of pharmaceutical and medical information). %

We believe that this result warrants new research on how to select the optimal number of modules, and on how to identify their best combination.

\setlength{\tabcolsep}{6pt}
\begin{table}[t]
    \centering
    \small
    \begin{tabular}{lccc}
        \toprule
          \textbf{Model} & \textbf{Regression} & \textbf{PearsonC} &  \textbf{SpearmanC} \\
        \midrule
        \multirow{4}{*}{{\small\texttt{gpt2-base}}} & Mean Diff.  & 0.8247* & 0.8152*\\

        & Linear & 0.9472* & 0.9640* \\

        &Ridge   & 0.9472* & 0.9641* \\

        \midrule
        \multirow{4}{*}{{\small\texttt{deberta-base}}} & Mean Diff. & 0.6584* & 0.6142*\\

        & Linear  & 0.9127*	& 0.9151*\\

        & Ridge  & 0.9168*	& 0.9225*\\

        \midrule
        \multirow{4}{*}{{\small\texttt{gpt2-large}}} & Mean Diff. & 0.8630* & 0.6857* \\
         & Linear  & 0.9636* &	0.9526*\\
         & Ridge  & 0.9683* &	0.9577*\\
        
        \bottomrule
    \end{tabular}
    \caption{Results of our meta-regression (mean correlation scores (Pearson and Spearman) obtained via 10-fold cross-validation, *statistically significant at $\alpha < 0.05$).}
    \label{tab:metaregression}
\end{table}

\section{Related Work}
We cover the related literature concerning the topics of knowledge modularization and knowledge composition. For a thorough overview of modular deep learning, we refer to \citet{modularDeepL}. 
\paragraph{Modularizing Knowledge.} 
Famously, \citet{houlsby2019parameterefficient} proposed to use adapter layers~\citep{RebuffiBV17} as a more efficient alternative to full task-specific fine-tuning. Subsequently, researchers in NLP explored adapters for various purposes, e.g., domain adaptation~\citep[e.g.,][]{glavas-etal-2021-training, cooper-stickland-etal-2021-multilingual, hung-etal-2022-ds, malik-etal-2023-udapter}, bias mitigation~\citep[e.g.,][]{lauscher-etal-2021-sustainable-modular, holtermann-etal-2022-fair, talat2022back}, language adaptation~\citep[e.g.,][]{philip-etal-2020-monolingual,ustun-etal-2022-udapter}, and for the injection of various other types of knowledge, such as common sense~\citep{lauscher-etal-2020-common}, factual~\citep{k-adapter}, and sociodemographic knowledge~\citep{hung-etal-2023-demographic}. 

Similarly, much effort has been spent designing new adapter variants with the aim of further increasing their efficiency or effectiveness~\citep[e.g.,][]{pfeiffer-etal-2021-adapterfusion, MahabadiHR21,zeng-etal-2023-one}. Alternatives to adapters that support modularity include subnetworks~\citep{guo-etal-2021-parameter} obtained via sparse fine-tuning, prefix tuning~\citep{li-liang-2021-prefix}, and mixture-of-expert~\citep[MoE;][]{6797059} models. 

The latter, exemplified by Switch Transformers~\citep{switchTransformer}, integrate a learned gating mechanism to channel inputs to appropriate expert modules. Like other modularization techniques, MoEs have been studied extensively for a wide range of problems \citep[e.g.,][]{gshard, kudugunta-etal-2021-beyond-distillation, nllbteam2022language, ponti-etal-2023-combining}. %
Most relevant to us, they have also been used to modularize different types of domain knowledge \citep{guo-etal-2018-multi, zhong2023metadmoe}. %
In this context, recent studies have considered experts as entirely autonomous models, challenging prevailing efficiency paradigms \citep{demix, li2022branchtrainmerge, CBTM}.

\paragraph{Composing Knowledge.} The composition of knowledge modules can be conducted via optimizing additional parameters~\citep[e.g.,][]{pfeiffer-etal-2021-adapterfusion}, or in a zero-shot manner~\citep[e.g.,][]{adapterSoup}.
Falling under the first category of approaches, \citet{pfeiffer-etal-2021-adapterfusion} proposed the fusion of adapters based on weights obtained via learned attention matrices. The same mechanism has been adopted by \citet{lu-etal-2021-parameter-efficient}, dubbed knowledge controller. In a similar vein, \citet{emea} ensemble the output vectors of multiple language adapters and optimize the respective ensemble weights. \citet{wang-etal-2022-adamix} and \citet{muqeeth2023soft} compose MoE models by learning to route the input to the right modules. Most recently, \citet{frohmann2023scalearn} propose to directly learn scaling parameters for efficient knowledge composition in task transfer.

In this work, we are interested in zero-shot knowledge composition. In this realm, \citet[][]{adapterSoup} rely on weight space averaging and simple selection strategies.  \citet{li2022branchtrainmerge} and \citet{CBTM} %
compare ensembling and averaging for composing domain PLMs, relying on domain prior for selection. Until now, a unified view is missing. %

\section{Conclusion}
In this work, we proposed a unified framework providing an interoperable notion of zero-shot knowledge composition. Using our framework, we analyzed the effectiveness of different knowledge module selection, weighting, and combination strategies. We studied the problem of domain adaptation with adapters and showed, for instance, that ensembling generally yields better results than parameter averaging. %
Examining five different scoring strategies, we found that even simple approaches can deliver strong results. Our findings also suggest that the number of adapters selected is generally more important than the weights assigned to them. 
While we have chosen the popular scenario of zero-shot domain adaptation with adapter layers, we are convinced that our framework is applicable to many other problems and modularization techniques (e.g., MoEs, entire models).

Overall, we believe that our results will fuel future research in effective knowledge composition by providing a consolidated perspective on zero-shot module composition.

\section*{Reproducibility Statement}
The 31 domain datasets we used for training and testing our domain adapters are publicly available and commonly used in other domain adaptation research. This facilitates comparability of our results with previous and future approaches and fosters the reproducibility of our results. 

We describe all datasets and splits in Section \ref{sec:experimental_setup} and Appendix \ref{app:data_code}. Additionally, all models we used for the experiments are publicly available in the Huggingface library~\citep{wolf-etal-2020-transformers}. Information on adapter training and inference, including details about hyperparameter settings, initialization, and hardware can be found in Section \ref{sec:experimental_setup}. Additional information about frameworks and code bases used are listed in Appendix \ref{app:data_code}. Finally, we release our code publicly under the MIT License to ensure open access to the community.

\section*{Limitations}
Naturally, our work comes with a number of limitations. Most importantly, we conducted our experiments on the C4 dataset only. However, we strongly believe our main findings to hold also for other corpora designed for testing domain adaptation methods. Related to this aspect, our notion of domains follows the one employed in C4 and is restricted to source websites as domain representatives. Previous research has shown that this definition is not always sufficient to clearly delineate domain knowledge~\citep[e.g.,][]{CBTM}. Therefore, we advise practitioners to carefully choose the criteria for discriminating among domains that are most useful in their particular application scenario. Additionally, our validation relies primarily on perplexity as a measure for general NLU of PLMs. While perplexity provides a robust initial measure, it does not encapsulate all facets of language understanding and generation, and only serves as a proxy for the final downstream performance of the models. Last, we resorted to adapters as the, arguably, most popular modularization technique in our experiments. We did not test other modularization approaches (e.g., MoEs) due to the large number of additional experiments required and related environmental considerations.   However, we strongly believe that our framework is general enough to provide useful guidance for the composition of various types of knowledge modularization techniques proposed in the literature. %

\section*{Ethical Considerations}
We also like to point to the ethical aspects touched by our work. First, as the large body of previous work on bias measurement demonstrates, PLMs are prone to encode and propagate stereotypical and exclusive biases present in their training data~\citep[e.g.,][]{bolukbasi, blodgett-etal-2020-language}. The models we used in our experiments are not spared from this issue \citep{deberta_bias, narayanan-venkit-etal-2023-nationality}. We advise practitioners to use these models with the appropriate care and we refer to existing works \citep{bias_mitigation_liang, lauscher-etal-2021-sustainable-modular} for discussions on bias mitigation. 
Second, central to our work are environmental considerations: experimentation with deep learning models potentially entails large amounts of CO$_2$ emissions \citep{Strubell_Ganesh_McCallum_2020}. With our work, we hope to encourage further research on efficient NLP, in particular on modular learning and module composition, and, hence, to contribute to greener AI.

\section*{Acknowledgements}
The work of Carolin Holtermann and Anne Lauscher is funded under the Excellence Strategy of the German Federal Government and the States. The work of Markus Frohmann and Navid Rekabsaz is partially funded by the State of Upper Austria and the Federal Ministry of Education, Science, and Research, through the grant number LIT-2021-YOU-215. We thank all anonymous reviewers for their valuable feedback.

\bibliography{custom}

\begin{thebibliography}{61}
\expandafter\ifx\csname natexlab\endcsname\relax\def\natexlab#1{#1}\fi

\bibitem[{Ansell et~al.(2021)Ansell, Ponti, Pfeiffer, Ruder, Glava{\v{s}}, Vuli{\'c}, and Korhonen}]{ansell-etal-2021-mad-g}
Alan Ansell, Edoardo~Maria Ponti, Jonas Pfeiffer, Sebastian Ruder, Goran Glava{\v{s}}, Ivan Vuli{\'c}, and Anna Korhonen. 2021.
\newblock \href {https://doi.org/10.18653/v1/2021.findings-emnlp.410} {{MAD}-{G}: {M}ultilingual adapter generation for efficient cross-lingual transfer}.
\newblock In \emph{Findings of the Association for Computational Linguistics: EMNLP 2021}, pages 4762--4781, Punta Cana, Dominican Republic. Association for Computational Linguistics.

\bibitem[{Bapna and Firat(2019)}]{bapna-firat-2019-simple}
Ankur Bapna and Orhan Firat. 2019.
\newblock \href {https://doi.org/10.18653/v1/D19-1165} {Simple, scalable adaptation for neural machine translation}.
\newblock In \emph{Proceedings of the 2019 Conference on Empirical Methods in Natural Language Processing and the 9th International Joint Conference on Natural Language Processing (EMNLP-IJCNLP)}, pages 1538--1548, Hong Kong, China. Association for Computational Linguistics.

\bibitem[{Blodgett et~al.(2020)Blodgett, Barocas, Daum{\'e}~III, and Wallach}]{blodgett-etal-2020-language}
Su~Lin Blodgett, Solon Barocas, Hal Daum{\'e}~III, and Hanna Wallach. 2020.
\newblock \href {https://doi.org/10.18653/v1/2020.acl-main.485} {Language (technology) is power: A critical survey of {``}bias{''} in {NLP}}.
\newblock In \emph{Proceedings of the 58th Annual Meeting of the Association for Computational Linguistics}, pages 5454--5476, Online. Association for Computational Linguistics.

\bibitem[{Bolukbasi et~al.(2016)Bolukbasi, Chang, Zou, Saligrama, and Kalai}]{bolukbasi}
Tolga Bolukbasi, Kai{-}Wei Chang, James~Y. Zou, Venkatesh Saligrama, and Adam Kalai. 2016.
\newblock \href {http://arxiv.org/abs/1607.06520} {Man is to computer programmer as woman is to homemaker? debiasing word embeddings}.
\newblock \emph{CoRR}, abs/1607.06520.

\bibitem[{Brown et~al.(2020)Brown, Mann, Ryder, Subbiah, Kaplan, Dhariwal, Neelakantan, Shyam, Sastry, Askell, Agarwal, Herbert-Voss, Krueger, Henighan, Child, Ramesh, Ziegler, Wu, Winter, Hesse, Chen, Sigler, Litwin, Gray, Chess, Clark, Berner, McCandlish, Radford, Sutskever, and Amodei}]{NEURIPS2020_1457c0d6}
Tom Brown, Benjamin Mann, Nick Ryder, Melanie Subbiah, Jared~D Kaplan, Prafulla Dhariwal, Arvind Neelakantan, Pranav Shyam, Girish Sastry, Amanda Askell, Sandhini Agarwal, Ariel Herbert-Voss, Gretchen Krueger, Tom Henighan, Rewon Child, Aditya Ramesh, Daniel Ziegler, Jeffrey Wu, Clemens Winter, Chris Hesse, Mark Chen, Eric Sigler, Mateusz Litwin, Scott Gray, Benjamin Chess, Jack Clark, Christopher Berner, Sam McCandlish, Alec Radford, Ilya Sutskever, and Dario Amodei. 2020.
\newblock \href {https://proceedings.neurips.cc/paper_files/paper/2020/file/1457c0d6bfcb4967418bfb8ac142f64a-Paper.pdf} {Language models are few-shot learners}.
\newblock In \emph{Advances in Neural Information Processing Systems}, volume~33, pages 1877--1901. Curran Associates, Inc.

\bibitem[{Chronopoulou et~al.(2023)Chronopoulou, Peters, Fraser, and Dodge}]{adapterSoup}
Alexandra Chronopoulou, Matthew~E. Peters, Alexander Fraser, and Jesse Dodge. 2023.
\newblock \href {https://doi.org/10.48550/ARXIV.2302.07027} {Adaptersoup: Weight averaging to improve generalization of pretrained language models}.

\bibitem[{Cooper~Stickland et~al.(2021)Cooper~Stickland, Berard, and Nikoulina}]{cooper-stickland-etal-2021-multilingual}
Asa Cooper~Stickland, Alexandre Berard, and Vassilina Nikoulina. 2021.
\newblock \href {https://aclanthology.org/2021.wmt-1.64} {Multilingual domain adaptation for {NMT}: Decoupling language and domain information with adapters}.
\newblock In \emph{Proceedings of the Sixth Conference on Machine Translation}, pages 578--598, Online. Association for Computational Linguistics.

\bibitem[{Dodge et~al.(2021)Dodge, Sap, Marasovi{\'c}, Agnew, Ilharco, Groeneveld, Mitchell, and Gardner}]{Cleaned-C4}
Jesse Dodge, Maarten Sap, Ana Marasovi{\'c}, William Agnew, Gabriel Ilharco, Dirk Groeneveld, Margaret Mitchell, and Matt Gardner. 2021.
\newblock \href {https://doi.org/10.18653/v1/2021.emnlp-main.98} {Documenting large webtext corpora: A case study on the colossal clean crawled corpus}.
\newblock In \emph{Proceedings of the 2021 Conference on Empirical Methods in Natural Language Processing}, pages 1286--1305, Online and Punta Cana, Dominican Republic. Association for Computational Linguistics.

\bibitem[{Emelin et~al.(2022)Emelin, Bonadiman, Alqahtani, Zhang, and Mansour}]{emelin-etal-2022-injecting}
Denis Emelin, Daniele Bonadiman, Sawsan Alqahtani, Yi~Zhang, and Saab Mansour. 2022.
\newblock \href {https://doi.org/10.18653/v1/2022.emnlp-main.820} {Injecting domain knowledge in language models for task-oriented dialogue systems}.
\newblock In \emph{Proceedings of the 2022 Conference on Empirical Methods in Natural Language Processing}, pages 11962--11974, Abu Dhabi, United Arab Emirates. Association for Computational Linguistics.

\bibitem[{Fedus et~al.(2022)Fedus, Zoph, and Shazeer}]{switchTransformer}
William Fedus, Barret Zoph, and Noam Shazeer. 2022.
\newblock \href {http://jmlr.org/papers/v23/21-0998.html} {Switch transformers: Scaling to trillion parameter models with simple and efficient sparsity}.
\newblock \emph{Journal of Machine Learning Research}, 23(120):1--39.

\bibitem[{Frohmann et~al.(2023)Frohmann, Holtermann, Masoudian, Lauscher, and Rekabsaz}]{frohmann2023scalearn}
Markus Frohmann, Carolin Holtermann, Shahed Masoudian, Anne Lauscher, and Navid Rekabsaz. 2023.
\newblock Scalearn: Simple and highly parameter-efficient task transfer by learning to scale.
\newblock \emph{arXiv preprint arXiv:2310.01217}.

\bibitem[{Glava{\v{s}} et~al.(2021)Glava{\v{s}}, Ganesh, and Somasundaran}]{glavas-etal-2021-training}
Goran Glava{\v{s}}, Ananya Ganesh, and Swapna Somasundaran. 2021.
\newblock \href {https://aclanthology.org/2021.bea-1.11} {Training and domain adaptation for supervised text segmentation}.
\newblock In \emph{Proceedings of the 16th Workshop on Innovative Use of NLP for Building Educational Applications}, pages 110--116, Online. Association for Computational Linguistics.

\bibitem[{Guo et~al.(2021)Guo, Rush, and Kim}]{guo-etal-2021-parameter}
Demi Guo, Alexander Rush, and Yoon Kim. 2021.
\newblock \href {https://doi.org/10.18653/v1/2021.acl-long.378} {Parameter-efficient transfer learning with diff pruning}.
\newblock In \emph{Proceedings of the 59th Annual Meeting of the Association for Computational Linguistics and the 11th International Joint Conference on Natural Language Processing (Volume 1: Long Papers)}, pages 4884--4896, Online. Association for Computational Linguistics.

\bibitem[{Guo et~al.(2018)Guo, Shah, and Barzilay}]{guo-etal-2018-multi}
Jiang Guo, Darsh Shah, and Regina Barzilay. 2018.
\newblock \href {https://doi.org/10.18653/v1/D18-1498} {Multi-source domain adaptation with mixture of experts}.
\newblock In \emph{Proceedings of the 2018 Conference on Empirical Methods in Natural Language Processing}, pages 4694--4703, Brussels, Belgium. Association for Computational Linguistics.

\bibitem[{Gururangan et~al.(2022)Gururangan, Lewis, Holtzman, Smith, and Zettlemoyer}]{demix}
Suchin Gururangan, Mike Lewis, Ari Holtzman, Noah~A. Smith, and Luke Zettlemoyer. 2022.
\newblock \href {https://doi.org/10.18653/v1/2022.naacl-main.407} {{DEM}ix layers: Disentangling domains for modular language modeling}.
\newblock In \emph{Proceedings of the 2022 Conference of the North American Chapter of the Association for Computational Linguistics: Human Language Technologies}, pages 5557--5576, Seattle, United States. Association for Computational Linguistics.

\bibitem[{Gururangan et~al.(2023)Gururangan, Li, Lewis, Shi, Althoff, Smith, and Zettlemoyer}]{CBTM}
Suchin Gururangan, Margaret Li, Mike Lewis, Weijia Shi, Tim Althoff, Noah~A. Smith, and Luke Zettlemoyer. 2023.
\newblock \href {http://arxiv.org/abs/2303.14177} {Scaling expert language models with unsupervised domain discovery}.

\bibitem[{He et~al.(2021)He, Liu, Gao, and Chen}]{deberta}
Pengcheng He, Xiaodong Liu, Jianfeng Gao, and Weizhu Chen. 2021.
\newblock \href {https://openreview.net/forum?id=XPZIaotutsD} {Deberta: decoding-enhanced bert with disentangled attention}.
\newblock In \emph{9th International Conference on Learning Representations, {ICLR} 2021, Virtual Event, Austria, May 3-7, 2021}. OpenReview.net.

\bibitem[{Hershcovich et~al.(2022)Hershcovich, Webersinke, Kraus, Bingler, and Leippold}]{hershcovich-etal-2022-towards}
Daniel Hershcovich, Nicolas Webersinke, Mathias Kraus, Julia Bingler, and Markus Leippold. 2022.
\newblock \href {https://doi.org/10.18653/v1/2022.emnlp-main.159} {Towards climate awareness in {NLP} research}.
\newblock In \emph{Proceedings of the 2022 Conference on Empirical Methods in Natural Language Processing}, pages 2480--2494, Abu Dhabi, United Arab Emirates. Association for Computational Linguistics.

\bibitem[{Holtermann et~al.(2022)Holtermann, Lauscher, and Ponzetto}]{holtermann-etal-2022-fair}
Carolin Holtermann, Anne Lauscher, and Simone Ponzetto. 2022.
\newblock \href {https://doi.org/10.18653/v1/2022.acl-long.541} {Fair and argumentative language modeling for computational argumentation}.
\newblock In \emph{Proceedings of the 60th Annual Meeting of the Association for Computational Linguistics (Volume 1: Long Papers)}, pages 7841--7861, Dublin, Ireland. Association for Computational Linguistics.

\bibitem[{Houlsby et~al.(2019)Houlsby, Giurgiu, Jastrzebski, Morrone, de~Laroussilhe, Gesmundo, Attariyan, and Gelly}]{houlsby2019parameterefficient}
Neil Houlsby, Andrei Giurgiu, Stanislaw Jastrzebski, Bruna Morrone, Quentin de~Laroussilhe, Andrea Gesmundo, Mona Attariyan, and Sylvain Gelly. 2019.
\newblock \href {http://arxiv.org/abs/1902.00751} {Parameter-efficient transfer learning for nlp}.

\bibitem[{Hung et~al.(2023)Hung, Lauscher, Hovy, Ponzetto, and Glava{\v{s}}}]{hung-etal-2023-demographic}
Chia-Chien Hung, Anne Lauscher, Dirk Hovy, Simone~Paolo Ponzetto, and Goran Glava{\v{s}}. 2023.
\newblock \href {https://doi.org/10.18653/v1/2023.findings-eacl.116} {Can demographic factors improve text classification? revisiting demographic adaptation in the age of transformers}.
\newblock In \emph{Findings of the Association for Computational Linguistics: EACL 2023}, pages 1565--1580, Dubrovnik, Croatia. Association for Computational Linguistics.

\bibitem[{Hung et~al.(2022)Hung, Lauscher, Ponzetto, and Glava{\v{s}}}]{hung-etal-2022-ds}
Chia-Chien Hung, Anne Lauscher, Simone Ponzetto, and Goran Glava{\v{s}}. 2022.
\newblock \href {https://doi.org/10.18653/v1/2022.findings-acl.72} {{DS}-{TOD}: Efficient domain specialization for task-oriented dialog}.
\newblock In \emph{Findings of the Association for Computational Linguistics: ACL 2022}, pages 891--904, Dublin, Ireland. Association for Computational Linguistics.

\bibitem[{Jacobs et~al.(1991)Jacobs, Jordan, Nowlan, and Hinton}]{6797059}
Robert~A. Jacobs, Michael~I. Jordan, Steven~J. Nowlan, and Geoffrey~E. Hinton. 1991.
\newblock \href {https://doi.org/10.1162/neco.1991.3.1.79} {Adaptive mixtures of local experts}.
\newblock \emph{Neural Computation}, 3(1):79--87.

\bibitem[{Kingma and Ba(2015)}]{AdamW}
Diederik~P. Kingma and Jimmy Ba. 2015.
\newblock \href {http://arxiv.org/abs/1412.6980} {Adam: {A} method for stochastic optimization}.
\newblock In \emph{3rd International Conference on Learning Representations, {ICLR} 2015, San Diego, CA, USA, May 7-9, 2015, Conference Track Proceedings}.

\bibitem[{Kudugunta et~al.(2021)Kudugunta, Huang, Bapna, Krikun, Lepikhin, Luong, and Firat}]{kudugunta-etal-2021-beyond-distillation}
Sneha Kudugunta, Yanping Huang, Ankur Bapna, Maxim Krikun, Dmitry Lepikhin, Minh-Thang Luong, and Orhan Firat. 2021.
\newblock \href {https://doi.org/10.18653/v1/2021.findings-emnlp.304} {Beyond distillation: Task-level mixture-of-experts for efficient inference}.
\newblock In \emph{Findings of the Association for Computational Linguistics: EMNLP 2021}, pages 3577--3599, Punta Cana, Dominican Republic. Association for Computational Linguistics.

\bibitem[{Lauscher et~al.(2021)Lauscher, Lueken, and Glava{\v{s}}}]{lauscher-etal-2021-sustainable-modular}
Anne Lauscher, Tobias Lueken, and Goran Glava{\v{s}}. 2021.
\newblock \href {https://doi.org/10.18653/v1/2021.findings-emnlp.411} {Sustainable modular debiasing of language models}.
\newblock In \emph{Findings of the Association for Computational Linguistics: EMNLP 2021}, pages 4782--4797, Punta Cana, Dominican Republic. Association for Computational Linguistics.

\bibitem[{Lauscher et~al.(2020)Lauscher, Majewska, Ribeiro, Gurevych, Rozanov, and Glava{\v{s}}}]{lauscher-etal-2020-common}
Anne Lauscher, Olga Majewska, Leonardo F.~R. Ribeiro, Iryna Gurevych, Nikolai Rozanov, and Goran Glava{\v{s}}. 2020.
\newblock \href {https://doi.org/10.18653/v1/2020.deelio-1.5} {Common sense or world knowledge? investigating adapter-based knowledge injection into pretrained transformers}.
\newblock In \emph{Proceedings of Deep Learning Inside Out (DeeLIO): The First Workshop on Knowledge Extraction and Integration for Deep Learning Architectures}, pages 43--49, Online. Association for Computational Linguistics.

\bibitem[{Lepikhin et~al.(2021)Lepikhin, Lee, Xu, Chen, Firat, Huang, Krikun, Shazeer, and Chen}]{gshard}
Dmitry Lepikhin, HyoukJoong Lee, Yuanzhong Xu, Dehao Chen, Orhan Firat, Yanping Huang, Maxim Krikun, Noam Shazeer, and Zhifeng Chen. 2021.
\newblock \href {https://openreview.net/forum?id=qrwe7XHTmYb} {Gshard: Scaling giant models with conditional computation and automatic sharding}.
\newblock In \emph{9th International Conference on Learning Representations, {ICLR} 2021, Virtual Event, Austria, May 3-7, 2021}. OpenReview.net.

\bibitem[{Lesota et~al.(2021)Lesota, Rekabsaz, Cohen, Grasserbauer, Eickhoff, and Schedl}]{uncertainty}
Oleg Lesota, Navid Rekabsaz, Daniel Cohen, Klaus~Antonius Grasserbauer, Carsten Eickhoff, and Markus Schedl. 2021.
\newblock \href {https://doi.org/10.1145/3471158.3472229} {A modern perspective on query likelihood with deep generative retrieval models}.
\newblock In \emph{Proceedings of the 2021 ACM SIGIR International Conference on Theory of Information Retrieval}, ICTIR '21, page 185–195, New York, NY, USA. Association for Computing Machinery.

\bibitem[{Li et~al.(2022)Li, Gururangan, Dettmers, Lewis, Althoff, Smith, and Zettlemoyer}]{li2022branchtrainmerge}
Margaret Li, Suchin Gururangan, Tim Dettmers, Mike Lewis, Tim Althoff, Noah~A. Smith, and Luke Zettlemoyer. 2022.
\newblock \href {http://arxiv.org/abs/2208.03306} {Branch-train-merge: Embarrassingly parallel training of expert language models}.

\bibitem[{Li and Liang(2021)}]{li-liang-2021-prefix}
Xiang~Lisa Li and Percy Liang. 2021.
\newblock \href {https://doi.org/10.18653/v1/2021.acl-long.353} {Prefix-tuning: Optimizing continuous prompts for generation}.
\newblock In \emph{Proceedings of the 59th Annual Meeting of the Association for Computational Linguistics and the 11th International Joint Conference on Natural Language Processing (Volume 1: Long Papers)}, pages 4582--4597, Online. Association for Computational Linguistics.

\bibitem[{Liang et~al.(2021)Liang, Wu, Morency, and Salakhutdinov}]{bias_mitigation_liang}
Paul~Pu Liang, Chiyu Wu, Louis-Philippe Morency, and Ruslan Salakhutdinov. 2021.
\newblock Towards understanding and mitigating social biases in language models.
\newblock In \emph{International Conference on Machine Learning}, pages 6565--6576. PMLR.

\bibitem[{Lu et~al.(2021)Lu, Dou, and Nguyen}]{lu-etal-2021-parameter-efficient}
Qiuhao Lu, Dejing Dou, and Thien~Huu Nguyen. 2021.
\newblock \href {https://doi.org/10.18653/v1/2021.findings-emnlp.325} {Parameter-efficient domain knowledge integration from multiple sources for biomedical pre-trained language models}.
\newblock In \emph{Findings of the Association for Computational Linguistics: EMNLP 2021}, pages 3855--3865, Punta Cana, Dominican Republic. Association for Computational Linguistics.

\bibitem[{Mahabadi et~al.(2021)Mahabadi, Henderson, and Ruder}]{MahabadiHR21}
Rabeeh~Karimi Mahabadi, James Henderson, and Sebastian Ruder. 2021.
\newblock \href {https://proceedings.neurips.cc/paper/2021/hash/081be9fdff07f3bc808f935906ef70c0-Abstract.html} {Compacter: Efficient low-rank hypercomplex adapter layers}.
\newblock In \emph{Advances in Neural Information Processing Systems 34: Annual Conference on Neural Information Processing Systems 2021, NeurIPS 2021, December 6-14, 2021, virtual}, pages 1022--1035.

\bibitem[{Malik et~al.(2023)Malik, Ramesh~Kashyap, Kan, and Poria}]{malik-etal-2023-udapter}
Bhavitvya Malik, Abhinav Ramesh~Kashyap, Min-Yen Kan, and Soujanya Poria. 2023.
\newblock \href {https://doi.org/10.18653/v1/2023.eacl-main.165} {{UDAPTER} - efficient domain adaptation using adapters}.
\newblock In \emph{Proceedings of the 17th Conference of the European Chapter of the Association for Computational Linguistics}, pages 2249--2263, Dubrovnik, Croatia. Association for Computational Linguistics.

\bibitem[{Muqeeth et~al.(2023)Muqeeth, Liu, and Raffel}]{muqeeth2023soft}
Mohammed Muqeeth, Haokun Liu, and Colin Raffel. 2023.
\newblock Soft merging of experts with adaptive routing.
\newblock \emph{arXiv preprint arXiv:2306.03745}.

\bibitem[{Narayanan~Venkit et~al.(2023)Narayanan~Venkit, Gautam, Panchanadikar, Huang, and Wilson}]{narayanan-venkit-etal-2023-nationality}
Pranav Narayanan~Venkit, Sanjana Gautam, Ruchi Panchanadikar, Ting-Hao Huang, and Shomir Wilson. 2023.
\newblock \href {https://doi.org/10.18653/v1/2023.eacl-main.9} {Nationality bias in text generation}.
\newblock In \emph{Proceedings of the 17th Conference of the European Chapter of the Association for Computational Linguistics}, pages 116--122, Dubrovnik, Croatia. Association for Computational Linguistics.

\bibitem[{Pfeiffer et~al.(2021)Pfeiffer, Kamath, R{\"u}ckl{\'e}, Cho, and Gurevych}]{pfeiffer-etal-2021-adapterfusion}
Jonas Pfeiffer, Aishwarya Kamath, Andreas R{\"u}ckl{\'e}, Kyunghyun Cho, and Iryna Gurevych. 2021.
\newblock \href {https://doi.org/10.18653/v1/2021.eacl-main.39} {{A}dapter{F}usion: Non-destructive task composition for transfer learning}.
\newblock In \emph{Proceedings of the 16th Conference of the European Chapter of the Association for Computational Linguistics: Main Volume}, pages 487--503, Online. Association for Computational Linguistics.

\bibitem[{Pfeiffer et~al.(2023)Pfeiffer, Ruder, Vulic, and Ponti}]{modularDeepL}
Jonas Pfeiffer, Sebastian Ruder, Ivan Vulic, and Edoardo~Maria Ponti. 2023.
\newblock \href {https://doi.org/10.48550/arXiv.2302.11529} {Modular deep learning}.
\newblock \emph{CoRR}, abs/2302.11529.

\bibitem[{Philip et~al.(2020)Philip, Berard, Gall{\'e}, and Besacier}]{philip-etal-2020-monolingual}
Jerin Philip, Alexandre Berard, Matthias Gall{\'e}, and Laurent Besacier. 2020.
\newblock \href {https://doi.org/10.18653/v1/2020.emnlp-main.361} {Monolingual adapters for zero-shot neural machine translation}.
\newblock In \emph{Proceedings of the 2020 Conference on Empirical Methods in Natural Language Processing (EMNLP)}, pages 4465--4470, Online. Association for Computational Linguistics.

\bibitem[{Plank(2016)}]{plank}
Barbara Plank. 2016.
\newblock \href {https://arxiv.org/abs/1608.07836} {What to do about non-standard (or non-canonical) language in {NLP}}.
\newblock \emph{ArXiv preprint}, abs/1608.07836.

\bibitem[{Ponti et~al.(2023)Ponti, Sordoni, Bengio, and Reddy}]{ponti-etal-2023-combining}
Edoardo~Maria Ponti, Alessandro Sordoni, Yoshua Bengio, and Siva Reddy. 2023.
\newblock \href {https://aclanthology.org/2023.eacl-main.49} {Combining parameter-efficient modules for task-level generalisation}.
\newblock In \emph{Proceedings of the 17th Conference of the European Chapter of the Association for Computational Linguistics}, pages 687--702, Dubrovnik, Croatia. Association for Computational Linguistics.

\bibitem[{Radford et~al.(2019)Radford, Wu, Child, Luan, Amodei, and Sutskever}]{GPT2}
Alec Radford, Jeffrey Wu, Rewon Child, David Luan, Dario Amodei, and Ilya Sutskever. 2019.
\newblock Language models are unsupervised multitask learners.
\newblock Technical report, OpenAI.

\bibitem[{Raffel et~al.(2020)Raffel, Shazeer, Roberts, Lee, Narang, Matena, Zhou, Li, and Liu}]{C4}
Colin Raffel, Noam Shazeer, Adam Roberts, Katherine Lee, Sharan Narang, Michael Matena, Yanqi Zhou, Wei Li, and Peter~J. Liu. 2020.
\newblock \href {http://jmlr.org/papers/v21/20-074.html} {Exploring the limits of transfer learning with a unified text-to-text transformer}.
\newblock \emph{J. Mach. Learn. Res.}, 21:140:1--140:67.

\bibitem[{Rebuffi et~al.(2017)Rebuffi, Bilen, and Vedaldi}]{RebuffiBV17}
Sylvestre{-}Alvise Rebuffi, Hakan Bilen, and Andrea Vedaldi. 2017.
\newblock \href {https://proceedings.neurips.cc/paper/2017/hash/e7b24b112a44fdd9ee93bdf998c6ca0e-Abstract.html} {Learning multiple visual domains with residual adapters}.
\newblock In \emph{Advances in Neural Information Processing Systems 30: Annual Conference on Neural Information Processing Systems 2017, December 4-9, 2017, Long Beach, CA, {USA}}, pages 506--516.

\bibitem[{Reimers and Gurevych(2019)}]{reimers-gurevych-2019-sentence}
Nils Reimers and Iryna Gurevych. 2019.
\newblock \href {https://doi.org/10.18653/v1/D19-1410} {Sentence-{BERT}: Sentence embeddings using {S}iamese {BERT}-networks}.
\newblock In \emph{Proceedings of the 2019 Conference on Empirical Methods in Natural Language Processing and the 9th International Joint Conference on Natural Language Processing (EMNLP-IJCNLP)}, pages 3982--3992, Hong Kong, China. Association for Computational Linguistics.

\bibitem[{Saunders(2021)}]{domain-adaptation}
Danielle Saunders. 2021.
\newblock \href {https://arxiv.org/abs/2104.06951} {Domain adaptation and multi-domain adaptation for neural machine translation: {A} survey}.
\newblock \emph{ArXiv preprint}, abs/2104.06951.

\bibitem[{Strubell et~al.(2020)Strubell, Ganesh, and McCallum}]{Strubell_Ganesh_McCallum_2020}
Emma Strubell, Ananya Ganesh, and Andrew McCallum. 2020.
\newblock \href {https://doi.org/10.1609/aaai.v34i09.7123} {Energy and policy considerations for modern deep learning research}.
\newblock \emph{Proceedings of the AAAI Conference on Artificial Intelligence}, 34(09):13693--13696.

\bibitem[{Sun et~al.(2020)Sun, Shao, Li, Liu, Yan, Qiu, and Huang}]{Sun_Shao_Li_Liu_Yan_Qiu_Huang_2020}
Tianxiang Sun, Yunfan Shao, Xiaonan Li, Pengfei Liu, Hang Yan, Xipeng Qiu, and Xuanjing Huang. 2020.
\newblock \href {https://doi.org/10.1609/aaai.v34i05.6424} {Learning sparse sharing architectures for multiple tasks}.
\newblock \emph{Proceedings of the AAAI Conference on Artificial Intelligence}, 34(05):8936--8943.

\bibitem[{Tal et~al.(2022)Tal, Magar, and Schwartz}]{deberta_bias}
Yarden Tal, Inbal Magar, and Roy Schwartz. 2022.
\newblock \href {https://doi.org/10.18653/v1/2022.gebnlp-1.13} {Fewer errors, but more stereotypes? the effect of model size on gender bias}.
\newblock In \emph{Proceedings of the 4th Workshop on Gender Bias in Natural Language Processing (GeBNLP)}, pages 112--120, Seattle, Washington. Association for Computational Linguistics.

\bibitem[{Talat and Lauscher(2022)}]{talat2022back}
Zeerak Talat and Anne Lauscher. 2022.
\newblock Back to the future: On potential histories in nlp.
\newblock \emph{arXiv preprint arXiv:2210.06245}.

\bibitem[{Team et~al.(2022)Team, Costa-jussà, Cross, Çelebi, Elbayad, Heafield, Heffernan, Kalbassi, Lam, Licht, Maillard, Sun, Wang, Wenzek, Youngblood, Akula, Barrault, Gonzalez, Hansanti, Hoffman, Jarrett, Sadagopan, Rowe, Spruit, Tran, Andrews, Ayan, Bhosale, Edunov, Fan, Gao, Goswami, Guzmán, Koehn, Mourachko, Ropers, Saleem, Schwenk, and Wang}]{nllbteam2022language}
NLLB Team, Marta~R. Costa-jussà, James Cross, Onur Çelebi, Maha Elbayad, Kenneth Heafield, Kevin Heffernan, Elahe Kalbassi, Janice Lam, Daniel Licht, Jean Maillard, Anna Sun, Skyler Wang, Guillaume Wenzek, Al~Youngblood, Bapi Akula, Loic Barrault, Gabriel~Mejia Gonzalez, Prangthip Hansanti, John Hoffman, Semarley Jarrett, Kaushik~Ram Sadagopan, Dirk Rowe, Shannon Spruit, Chau Tran, Pierre Andrews, Necip~Fazil Ayan, Shruti Bhosale, Sergey Edunov, Angela Fan, Cynthia Gao, Vedanuj Goswami, Francisco Guzmán, Philipp Koehn, Alexandre Mourachko, Christophe Ropers, Safiyyah Saleem, Holger Schwenk, and Jeff Wang. 2022.
\newblock \href {http://arxiv.org/abs/2207.04672} {No language left behind: Scaling human-centered machine translation}.

\bibitem[{Tenney et~al.(2019)Tenney, Das, and Pavlick}]{tenney-etal-2019-bert}
Ian Tenney, Dipanjan Das, and Ellie Pavlick. 2019.
\newblock \href {https://doi.org/10.18653/v1/P19-1452} {{BERT} rediscovers the classical {NLP} pipeline}.
\newblock In \emph{Proceedings of the 57th Annual Meeting of the Association for Computational Linguistics}, pages 4593--4601, Florence, Italy. Association for Computational Linguistics.

\bibitem[{{\"U}st{\"u}n et~al.(2022){\"U}st{\"u}n, Bisazza, Bouma, and van Noord}]{ustun-etal-2022-udapter}
Ahmet {\"U}st{\"u}n, Arianna Bisazza, Gosse Bouma, and Gertjan van Noord. 2022.
\newblock \href {https://doi.org/10.1162/coli_a_00443} {{UD}apter: Typology-based language adapters for multilingual dependency parsing and sequence labeling}.
\newblock \emph{Computational Linguistics}, 48(3):555--592.

\bibitem[{Wang et~al.(2021{\natexlab{a}})Wang, Tang, Duan, Wei, Huang, Ji, Cao, Jiang, and Zhou}]{k-adapter}
Ruize Wang, Duyu Tang, Nan Duan, Zhongyu Wei, Xuanjing Huang, Jianshu Ji, Guihong Cao, Daxin Jiang, and Ming Zhou. 2021{\natexlab{a}}.
\newblock \href {https://doi.org/10.18653/v1/2021.findings-acl.121} {K-adapter: Infusing knowledge into pre-trained models with adapters}.
\newblock In \emph{Findings of the Association for Computational Linguistics: {ACL/IJCNLP} 2021, Online Event, August 1-6, 2021}, volume {ACL/IJCNLP} 2021 of \emph{Findings of {ACL}}, pages 1405--1418. Association for Computational Linguistics.

\bibitem[{Wang et~al.(2021{\natexlab{b}})Wang, Tsvetkov, Ruder, and Neubig}]{emea}
Xinyi Wang, Yulia Tsvetkov, Sebastian Ruder, and Graham Neubig. 2021{\natexlab{b}}.
\newblock \href {https://doi.org/10.18653/v1/2021.findings-emnlp.63} {Efficient test time adapter ensembling for low-resource language varieties}.
\newblock In \emph{Findings of the Association for Computational Linguistics: EMNLP 2021}, pages 730--737, Punta Cana, Dominican Republic. Association for Computational Linguistics.

\bibitem[{Wang et~al.(2022)Wang, Agarwal, Mukherjee, Liu, Gao, Awadallah, and Gao}]{wang-etal-2022-adamix}
Yaqing Wang, Sahaj Agarwal, Subhabrata Mukherjee, Xiaodong Liu, Jing Gao, Ahmed~Hassan Awadallah, and Jianfeng Gao. 2022.
\newblock \href {https://aclanthology.org/2022.emnlp-main.388} {{A}da{M}ix: Mixture-of-adaptations for parameter-efficient model tuning}.
\newblock In \emph{Proceedings of the 2022 Conference on Empirical Methods in Natural Language Processing}, pages 5744--5760, Abu Dhabi, United Arab Emirates. Association for Computational Linguistics.

\bibitem[{Wolf et~al.(2020)Wolf, Debut, Sanh, Chaumond, Delangue, Moi, Cistac, Rault, Louf, Funtowicz, Davison, Shleifer, von Platen, Ma, Jernite, Plu, Xu, Le~Scao, Gugger, Drame, Lhoest, and Rush}]{wolf-etal-2020-transformers}
Thomas Wolf, Lysandre Debut, Victor Sanh, Julien Chaumond, Clement Delangue, Anthony Moi, Pierric Cistac, Tim Rault, Remi Louf, Morgan Funtowicz, Joe Davison, Sam Shleifer, Patrick von Platen, Clara Ma, Yacine Jernite, Julien Plu, Canwen Xu, Teven Le~Scao, Sylvain Gugger, Mariama Drame, Quentin Lhoest, and Alexander Rush. 2020.
\newblock \href {https://doi.org/10.18653/v1/2020.emnlp-demos.6} {Transformers: State-of-the-art natural language processing}.
\newblock In \emph{Proceedings of the 2020 Conference on Empirical Methods in Natural Language Processing: System Demonstrations}, pages 38--45, Online. Association for Computational Linguistics.

\bibitem[{Wortsman et~al.(2022)Wortsman, Ilharco, Gadre, Roelofs, Gontijo-Lopes, Morcos, Namkoong, Farhadi, Carmon, Kornblith, and Schmidt}]{wortsman2022model}
Mitchell Wortsman, Gabriel Ilharco, Samir~Yitzhak Gadre, Rebecca Roelofs, Raphael Gontijo-Lopes, Ari~S. Morcos, Hongseok Namkoong, Ali Farhadi, Yair Carmon, Simon Kornblith, and Ludwig Schmidt. 2022.
\newblock \href {http://arxiv.org/abs/2203.05482} {Model soups: averaging weights of multiple fine-tuned models improves accuracy without increasing inference time}.

\bibitem[{Zeng et~al.(2023)Zeng, Zhang, and Lu}]{zeng-etal-2023-one}
Guangtao Zeng, Peiyuan Zhang, and Wei Lu. 2023.
\newblock \href {https://aclanthology.org/2023.acl-long.418} {One network, many masks: Towards more parameter-efficient transfer learning}.
\newblock In \emph{Proceedings of the 61st Annual Meeting of the Association for Computational Linguistics (Volume 1: Long Papers)}, pages 7564--7580, Toronto, Canada. Association for Computational Linguistics.

\bibitem[{Zhong et~al.(2023)Zhong, Chi, Gu, Wang, Yu, and Tang}]{zhong2023metadmoe}
Tao Zhong, Zhixiang Chi, Li~Gu, Yang Wang, Yuanhao Yu, and Jin Tang. 2023.
\newblock \href {http://arxiv.org/abs/2210.03885} {Meta-dmoe: Adapting to domain shift by meta-distillation from mixture-of-experts}.

\end{thebibliography}

\clearpage
\newpage
\onecolumn
\section*{Appendix}
\appendix
\section{Link to Data, Models, Code Bases} \label{app:data_code}
In Table \ref{tab:artifacts}, we provide all information and links to the data, models, frameworks, and code bases we use in our work. All artifacts were used according to their intended use, as described in their licenses. As described in the main body of this manuscript, we are also releasing our code publicly (MIT License).

\setlength{\tabcolsep}{3pt}
\begin{table}[h]
\small
    \centering
    \begin{tabular}{clp{5.0cm}p{5.0cm}}
    \toprule
    Purpose & Name & URL & Details\\
    \midrule
        \multirow{7}{*}{Code Base} & Language Modeling MLM & \small\url{https://github.com/adapter-hub/adapter-transformers/blob/master/examples/pytorch/language-modeling/run_mlm.py} \\
         & Language Modeling CLM & \small\url{https://github.com/adapter-hub/adapter-transformers/blob/master/examples/pytorch/language-modeling/run_clm.py} \\
        \midrule
        \multirow{7}{*}{Models} & \texttt{gpt2-base} & \url{https://huggingface.co/gpt2} & 12-layers, 768-hidden, 12-heads, 117M parameters\\
         & \texttt{gpt2-large} & \url{https://huggingface.co/gpt2-large} & 36-layers, 1280-hidden, 20-heads, 774M parameters\\
          & \texttt{deberta-base} & \url{https://huggingface.co/microsoft/deberta-base} & 12-layers, 768-hidden, 12-heads\\
        & SentenceBert & \url{https://github.com/UKPLab/sentence-transformers} & Configuration: \texttt{all-mpnet-base-v2} \\
        \midrule
        \multirow{8}{*}{Frameworks} & nltk==3.7 & & We use NLTK for punctuation removal, stemming and tokenization before creating the TF-IDF vectors.\\
        & adapter-transformers==3.2.1 & &\\
        & huggingface-hub==0.13.4  && \\
        & torch==2.0.0 & & \\
        & torchaudio==2.0.1 & & \\
        & torchvision==0.15.1 & & \\
        & transformers==4.28.1 & & \\
        & datasets==2.11.0 && \\
        \midrule
        \multirow{6}{*}{Datasets} & C4 & \url{https://github.com/allenai/c4-documentation} & License: ODC-BY\\
        & yelp.com & \url{https://www.yelp.com/dataset}& Licence: \url{https://s3-media0.fl.yelpcdn.com/assets/srv0/engineering_pages/f64cb2d3efcc/assets/vendor/Dataset_User_Agreement.pdf}\\
        
        \bottomrule
    \end{tabular}
    \caption{Links and explanations to code bases, datasets, models and frameworks used in our work.}
    \label{tab:artifacts}
\end{table}

\twocolumn

\section{TF--IDF Equation}\label{sec:tfidf_equation}
We determine the TF--IDF scores by:

\begin{multline*}%
tf\-idf(t,d) = tf(t,d) * idf(t) \\
tf(t,d) = \frac{f_{t,d}}{\sum_{t'\in d} f_{t',d}} \\
idf(t) = log \left( \frac{1 + N}{1 + df(t)}+1 \right),
 \\
\end{multline*}
where $N$ is the total number of documents.

\section{Comparison of Combination Strategies}\label{sec:app_avg_ens}
\begin{figure}[h]
    \centering
    \includegraphics[width=\linewidth]{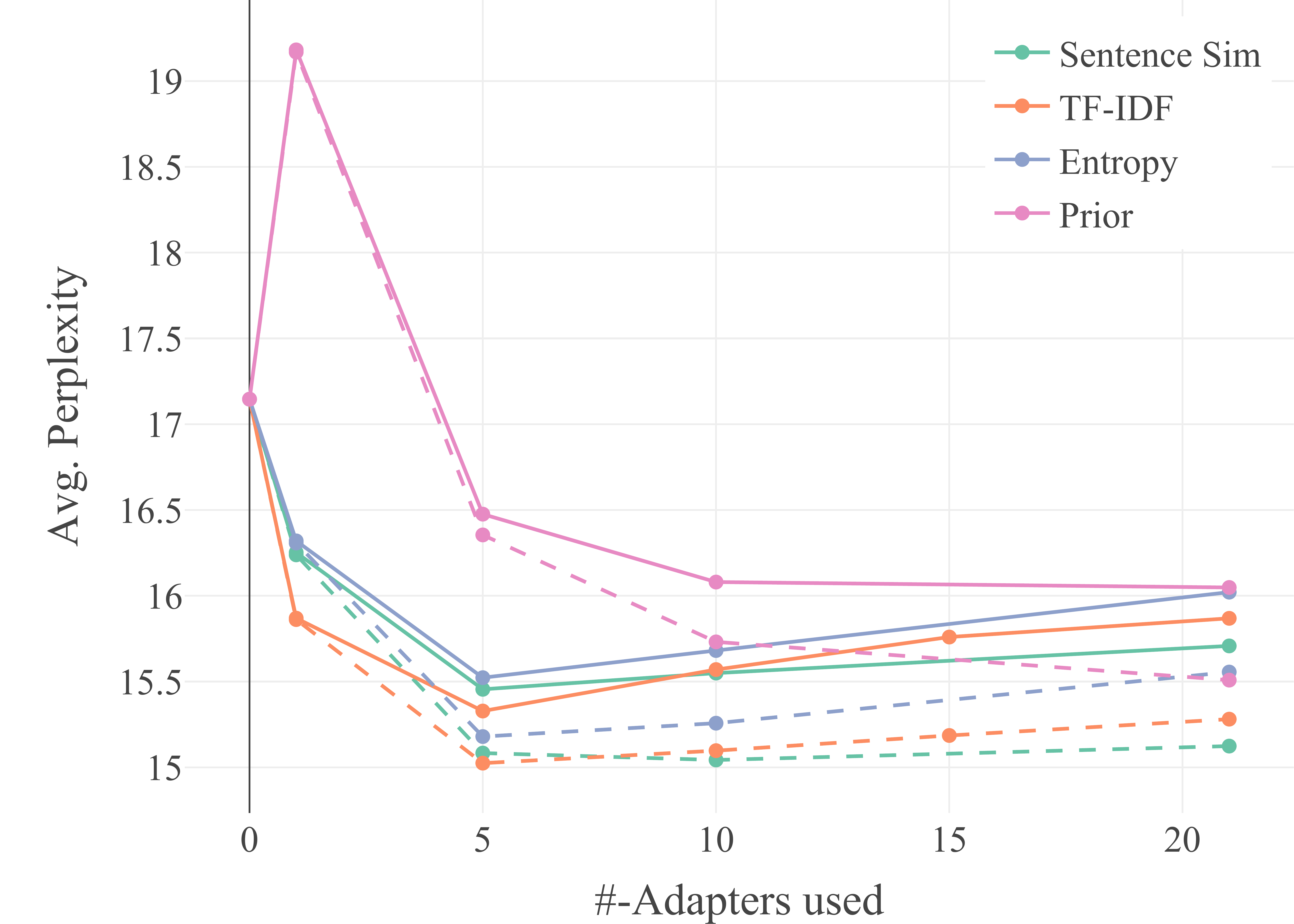}
    \caption{Comparison between Parameter Averaging (solid lines) and Ensembling (dashed lines) for {\small\texttt{gpt2-large}} over different numbers of top-$k$ adapters. We show the mean perplexity results when using each of our four scoring strategies (\textsc{SentSim}, \textsc{tf--idf}, \textsc{entropy}, \textsc{prior}) averaged across four runs.}
    \label{fig:gpt2_l_avg_ens}
\end{figure}

We evaluate the combination strategies for three different models. In Figure \ref{fig:gpt2_l_avg_ens}, we present the results for ensembling and parameter averaging for {\small\texttt{gpt2-large}}. Compared to the results for {\small\texttt{gpt2-base}} and {\small\texttt{deberta-base}}, which we showed in Figure \ref{fig:AVG_ENS_Comp}, we did not run the experiments for all values for $k$ between [0,10] because of the size of the model. However, we find very similar patterns in the variation of perplexity across the different strategies and number of adapters added as for {\small\texttt{gpt2-base}}. This reinforces the validity of our findings.

\begin{figure}[h!]
     \centering
     \begin{subfigure}[b]{0.9\columnwidth}
         \centering
         \includegraphics[width=\textwidth]{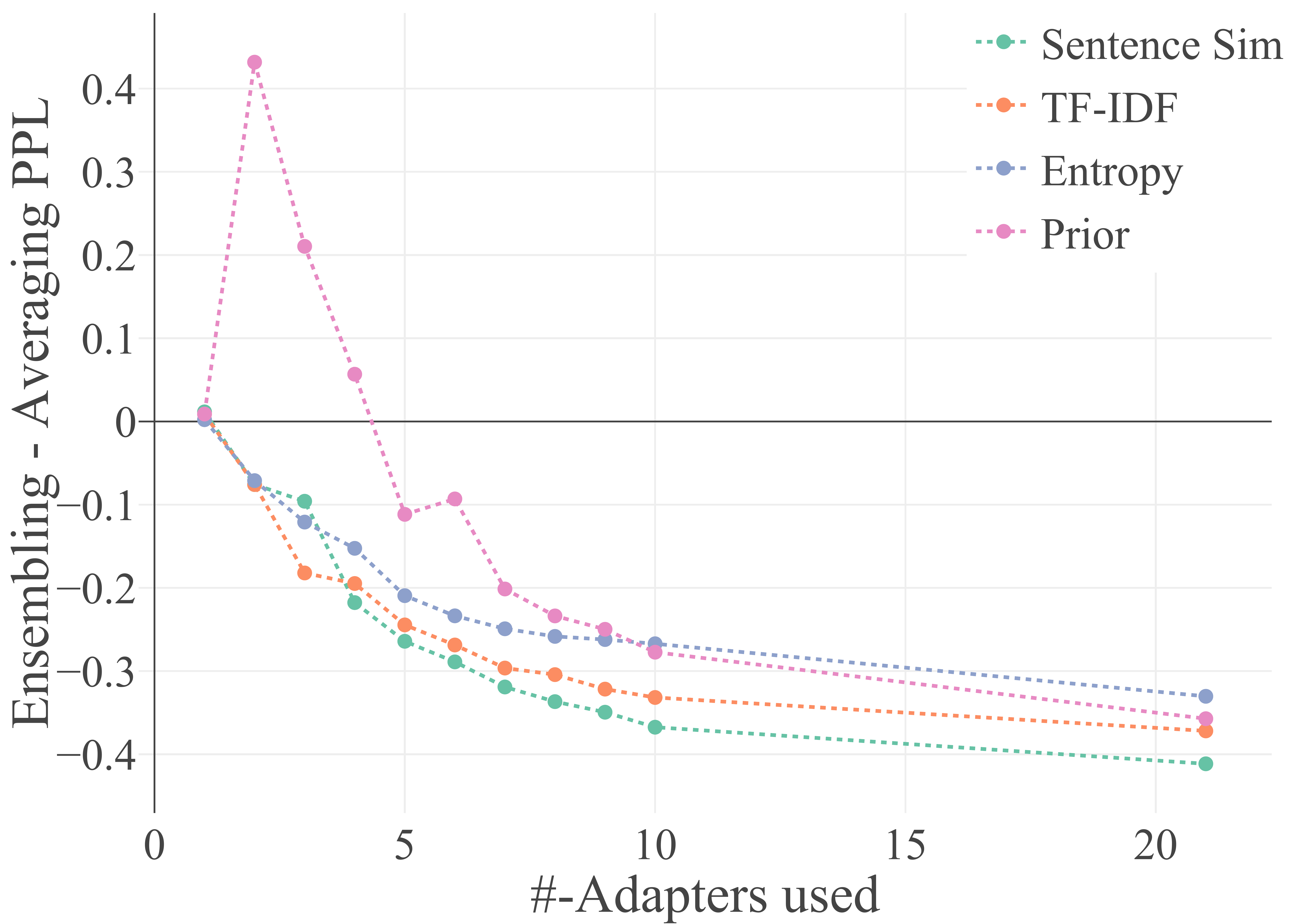}
         \caption{\small\texttt{gpt2-base}}
         \label{fig:diff_gpt}
     \end{subfigure}
     \hfill
     \begin{subfigure}[b]{0.9\columnwidth}
         \centering
         \includegraphics[width=\textwidth]{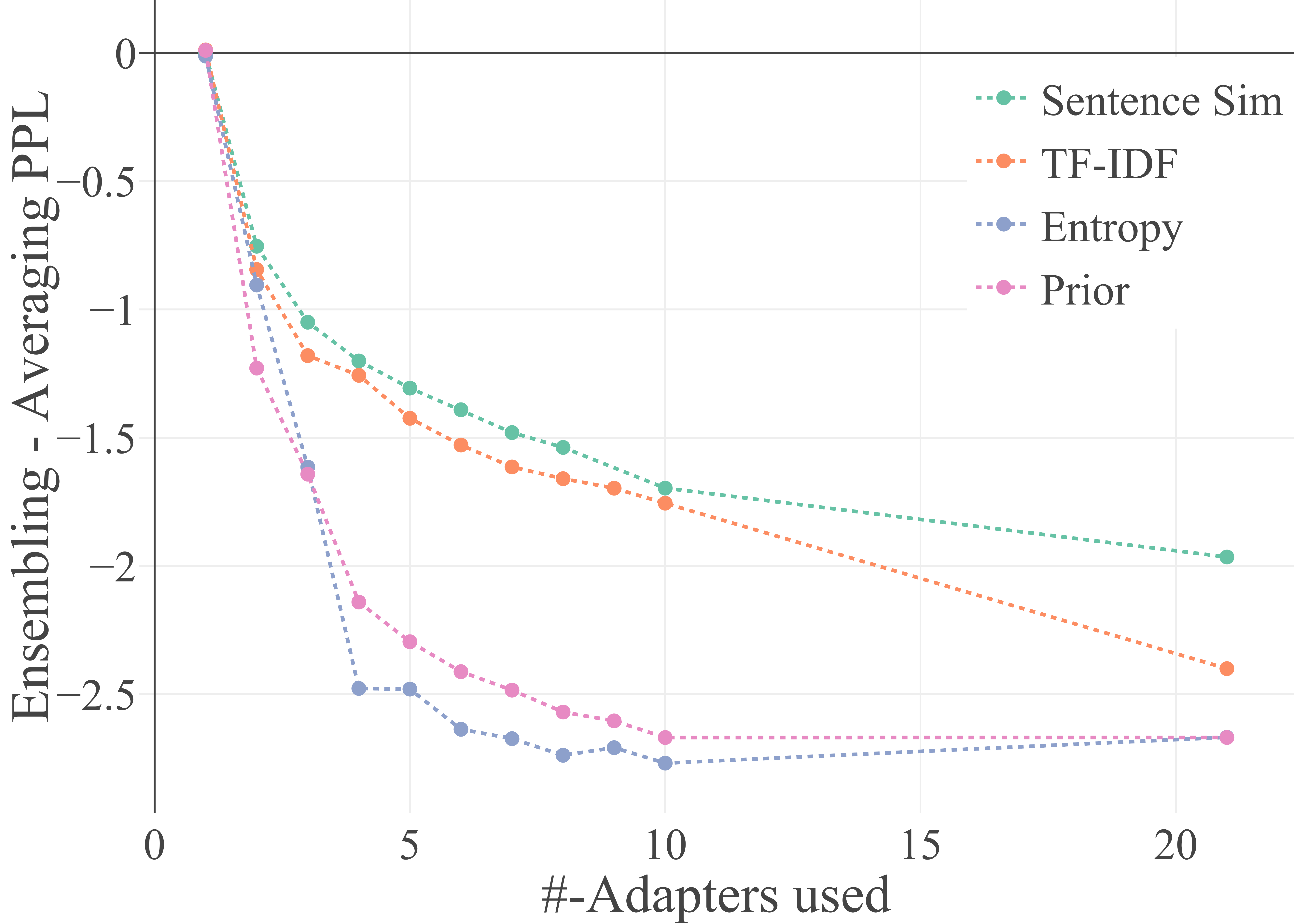}
         \caption{\small\texttt{deberta-base}}
         \label{fig:diff_deberta}
     \end{subfigure}
     \hfill
     \begin{subfigure}[b]{0.9\columnwidth}
         \centering
         \includegraphics[width=\textwidth]{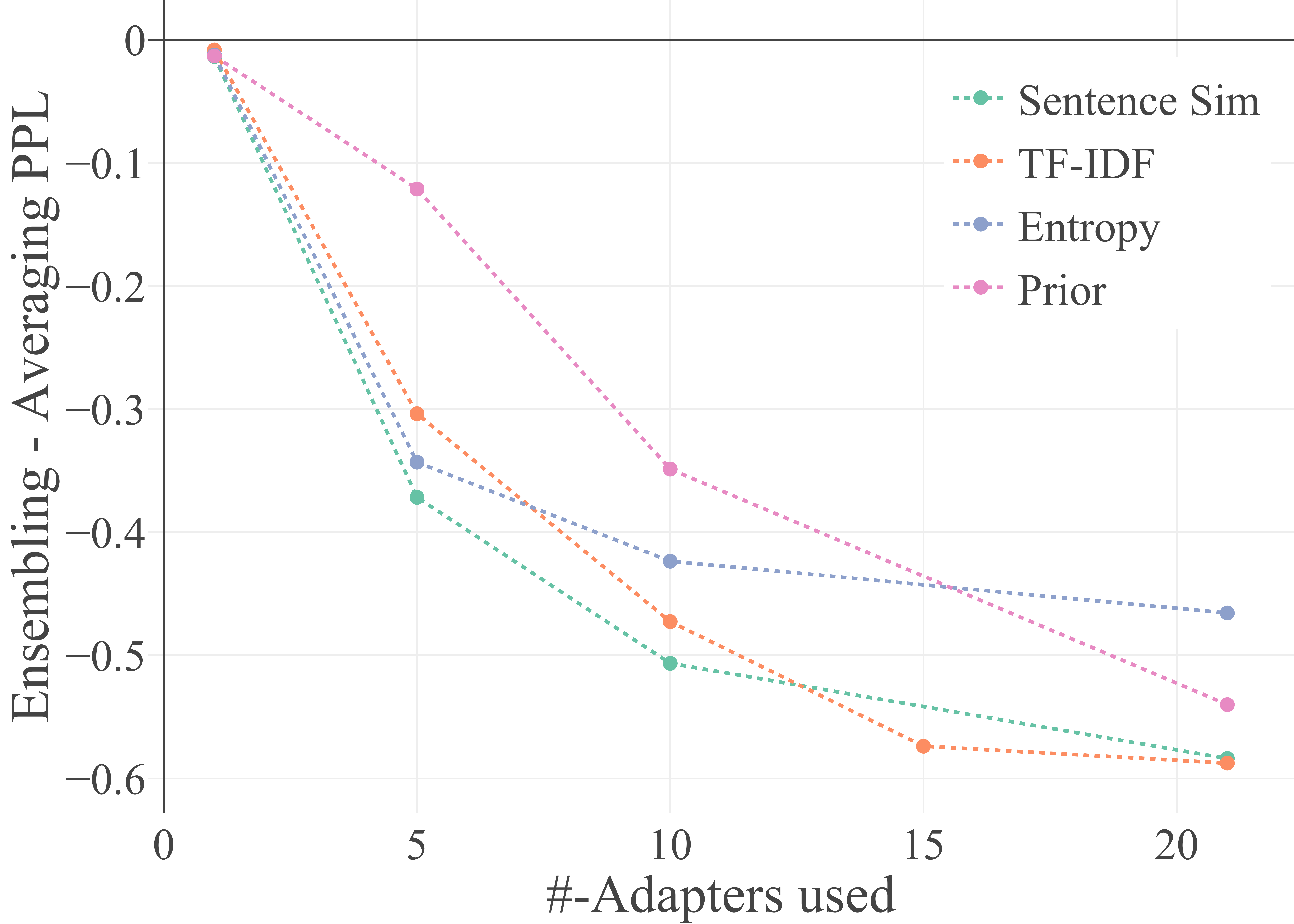}
         \caption{\small\texttt{gpt2-large}}
         \label{fig:diff_gpt2_large}
     \end{subfigure}
     \hfill
        \caption{Difference between Ensembling - Parameter Averaging over different numbers of top-$k$ adapters. We show the mean perplexity differences for (a) {\small\texttt{gpt2-base}}, and (b) {\small\texttt{deberta-base}} (c) {\small\texttt{gpt2-large}} when using each of our four scoring strategies (\textsc{SentSim}, \textsc{tf--idf}, \textsc{entropy}, \textsc{prior}) averaged across four runs.
        }
        \label{fig:diff_avg_ens}
\end{figure}

Figure \ref{fig:diff_avg_ens} additionally shows the perplexity difference between parameter averaging and ensembling for the different scoring strategies. A negative value indicates that ensembling provides lower perplexity values than parameter averaging. 

Interestingly, we can see the same tendency for all three models. With an increasing value of $k$, the difference between parameter averaging and ensembling increases as well, although this effect flattens for $k>10$. For {\small\texttt{deberta-base}}, this effect can be seen more strongly. Interestingly, while for {\small\texttt{deberta-base}}, the difference is larger for model-based approaches, we see an exact opposite effect for the GPT-models.

\clearpage

\section{Meta Regression}\label{sec:meta_reg_coefs}
We present the coefficients of linear regression for {\small\texttt{gpt2-base}}, {\small\texttt{deberta-base}} and {\small\texttt{gpt2-large}}. We do not include coefficients with an importance value between [-0.1, 0.1].

\begin{figure}[h]
    \centering
    \includegraphics[width=3.8cm]{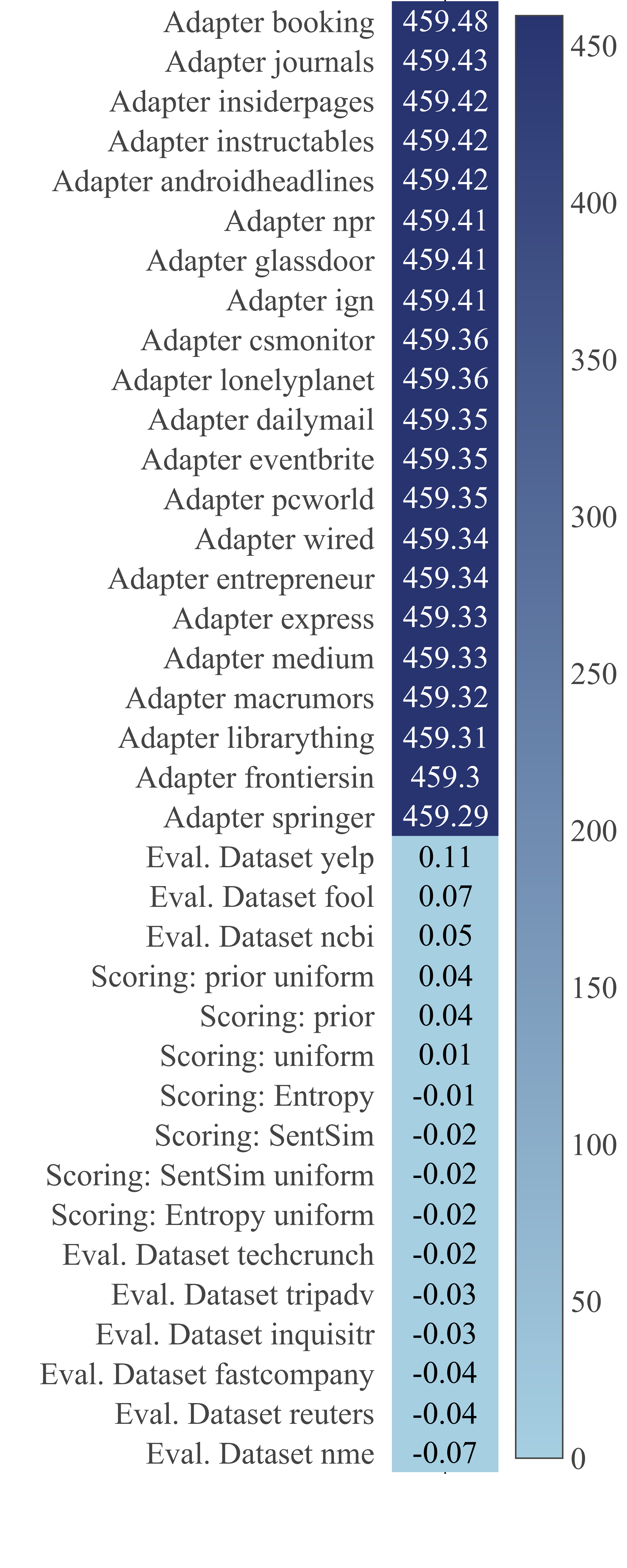}
    \caption{Heatmap of the coefficients of the Linear Regression for {\small\texttt{gpt2-base}}}
    \label{fig:heatmap_gpt2}
\end{figure}

\begin{figure}[h]
    \centering
    \includegraphics[width=3.5cm]{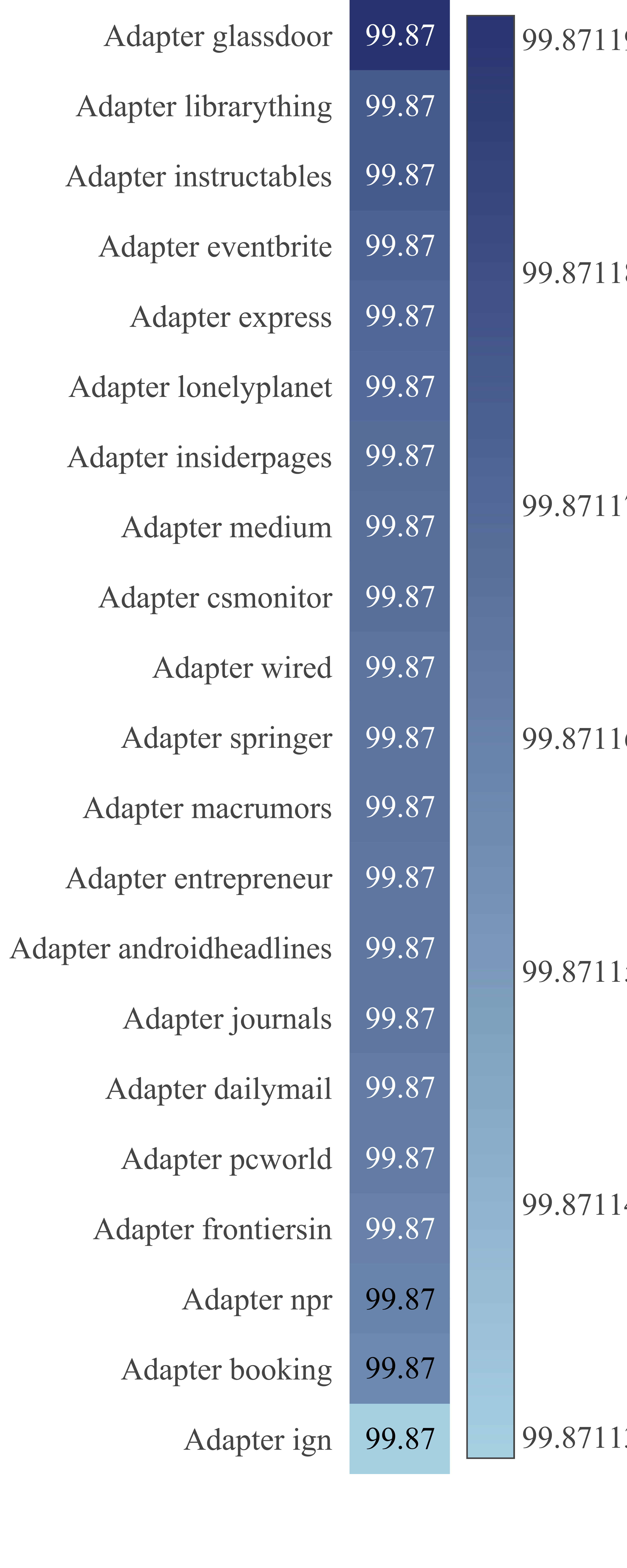}
    \caption{Heatmap of the coefficients of the Linear Regression for {\small\texttt{deberta-base}}}
    \label{fig:heatmap_deberta}
\end{figure}

\begin{figure}[h]
    \centering
    \includegraphics[width=3.5cm]{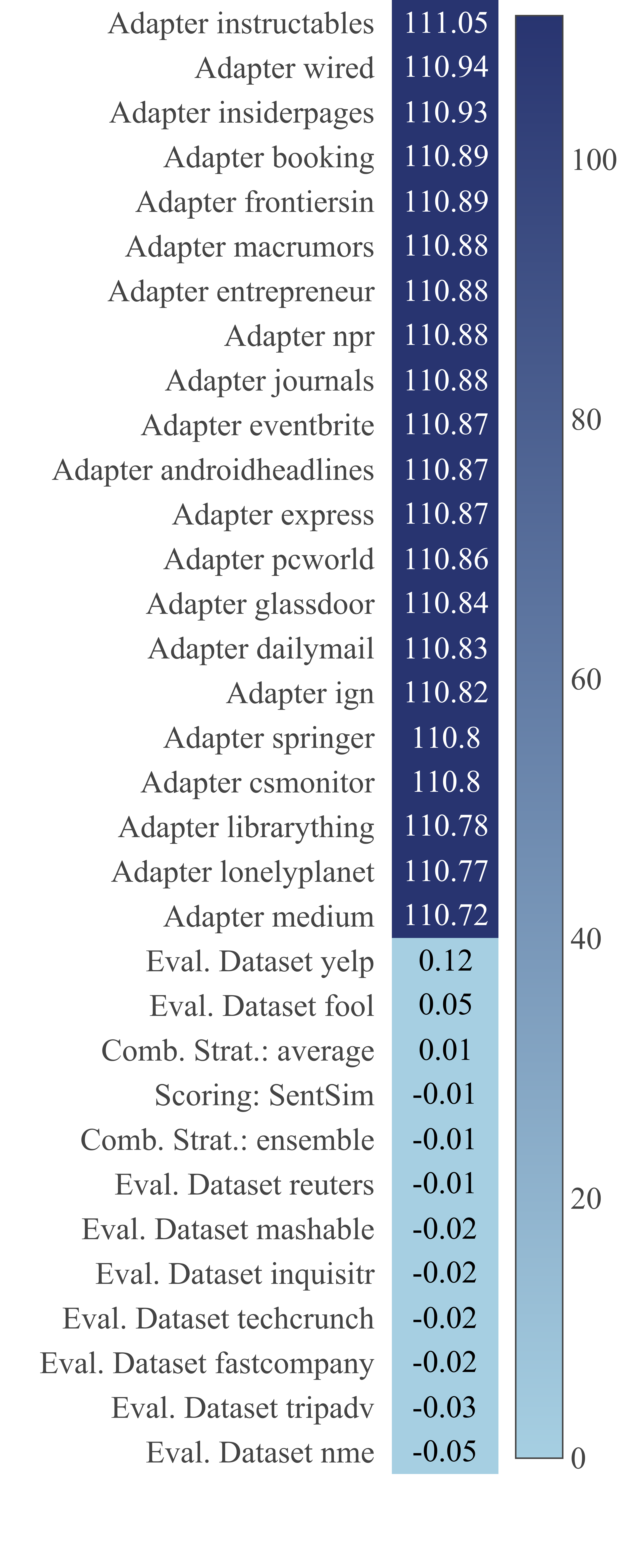}
    \caption{Heatmap of the coefficients of the Linear Regression for {\small\texttt{gpt2-large}}}
    \label{fig:heatmap_gpt2l}
\end{figure}

\clearpage

\onecolumn

\section{Further Evaluation of Adapter Scorings}\label{sec:weighted_unweighted_comp}

\begin{figure*}[h]
     \centering
     \begin{subfigure}[b]{0.3\textwidth}
         \centering
         \includegraphics[width=\textwidth]{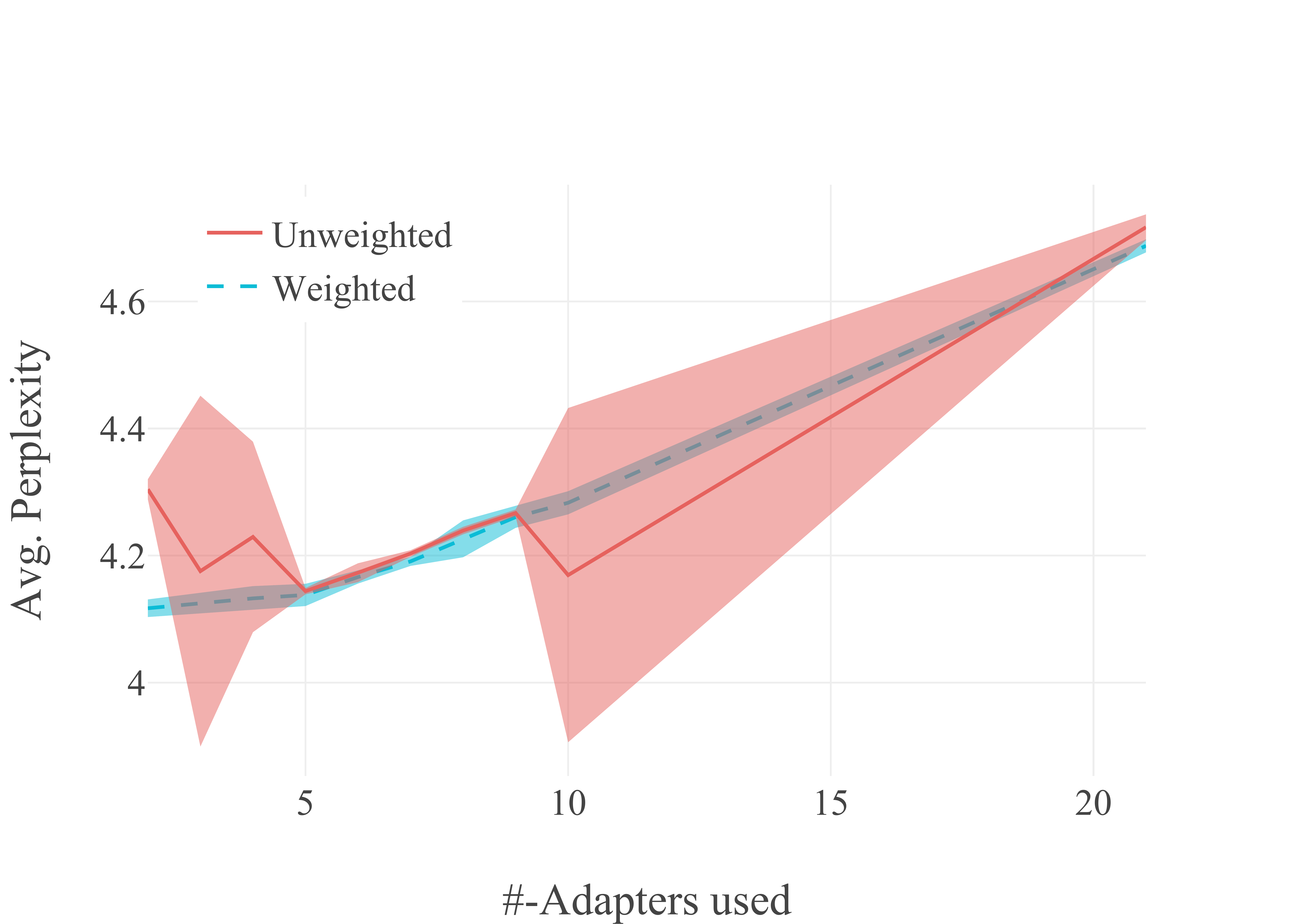}
         \caption{\textsc{entropy} - ensemble}
     \end{subfigure}
     \hfill
     \begin{subfigure}[b]{0.3\textwidth}
         \centering
         \includegraphics[width=\textwidth]{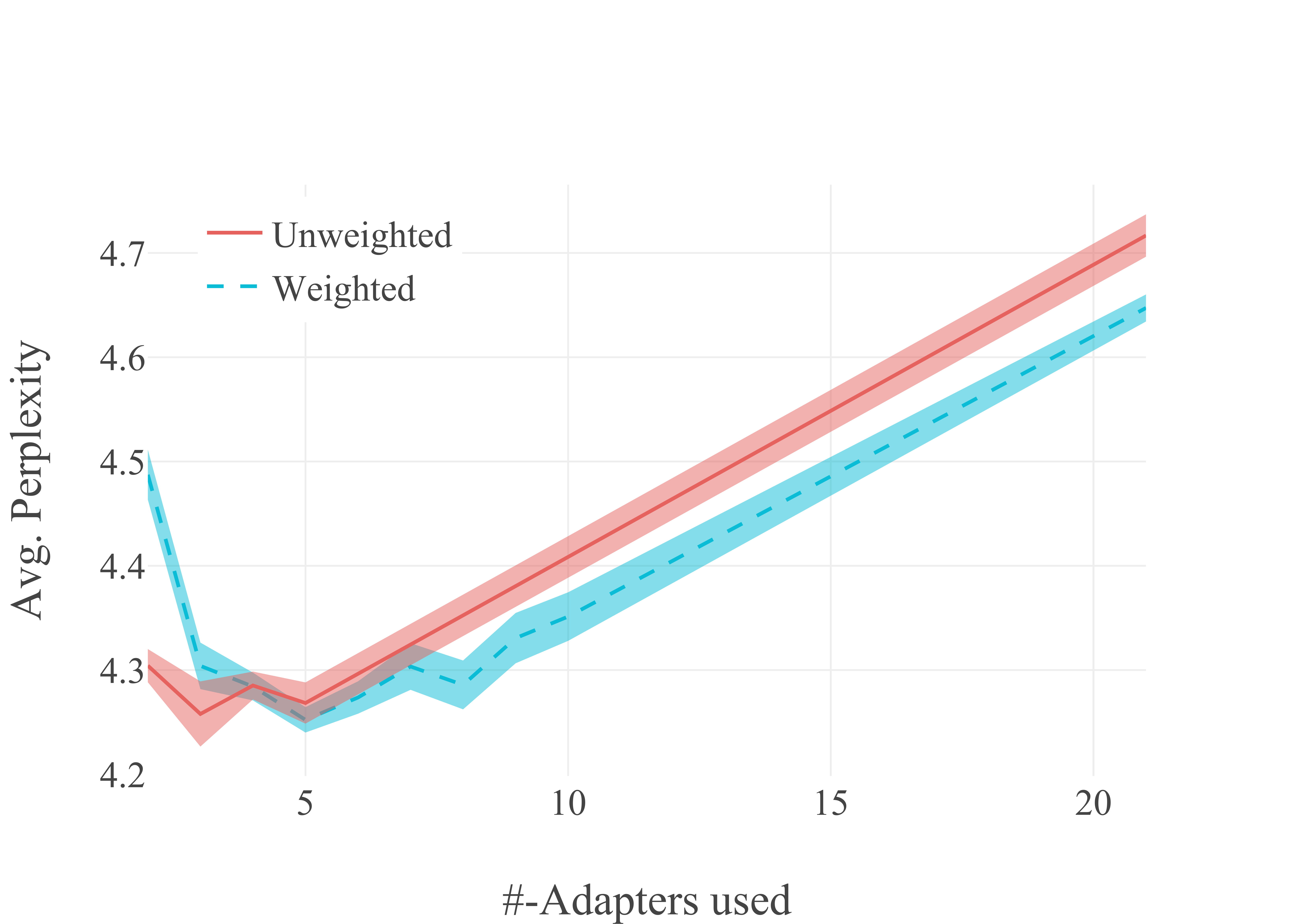}
         \caption{\textsc{prior} - ensemble}
     \end{subfigure}
     \hfill
     \begin{subfigure}[b]{0.3\textwidth}
         \centering
         \includegraphics[width=\textwidth]{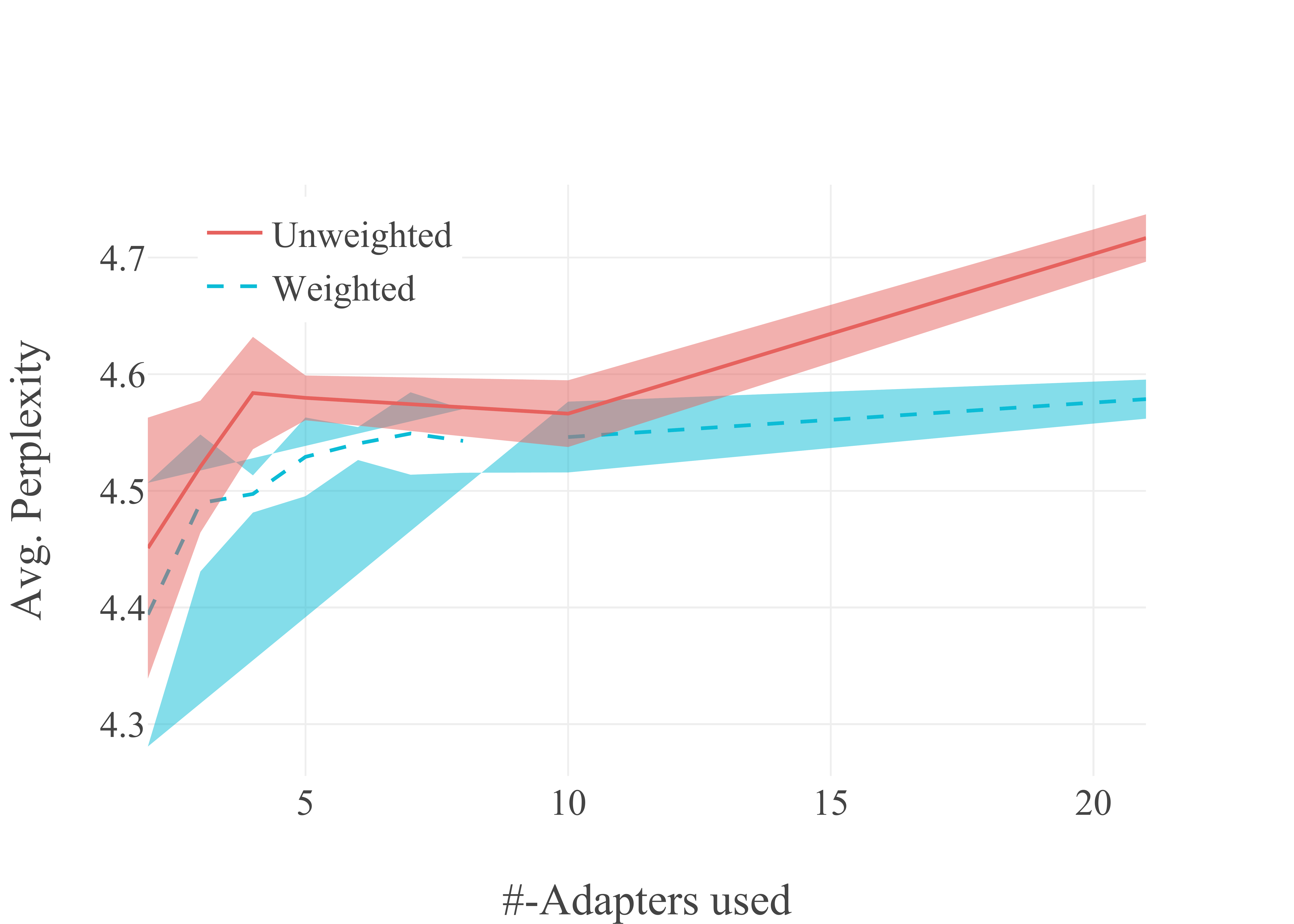}
         \caption{\textsc{SentSim} - ensemble}
     \end{subfigure}
     \hfill
     \begin{subfigure}[b]{0.3\textwidth}
         \centering
         \includegraphics[width=\textwidth]{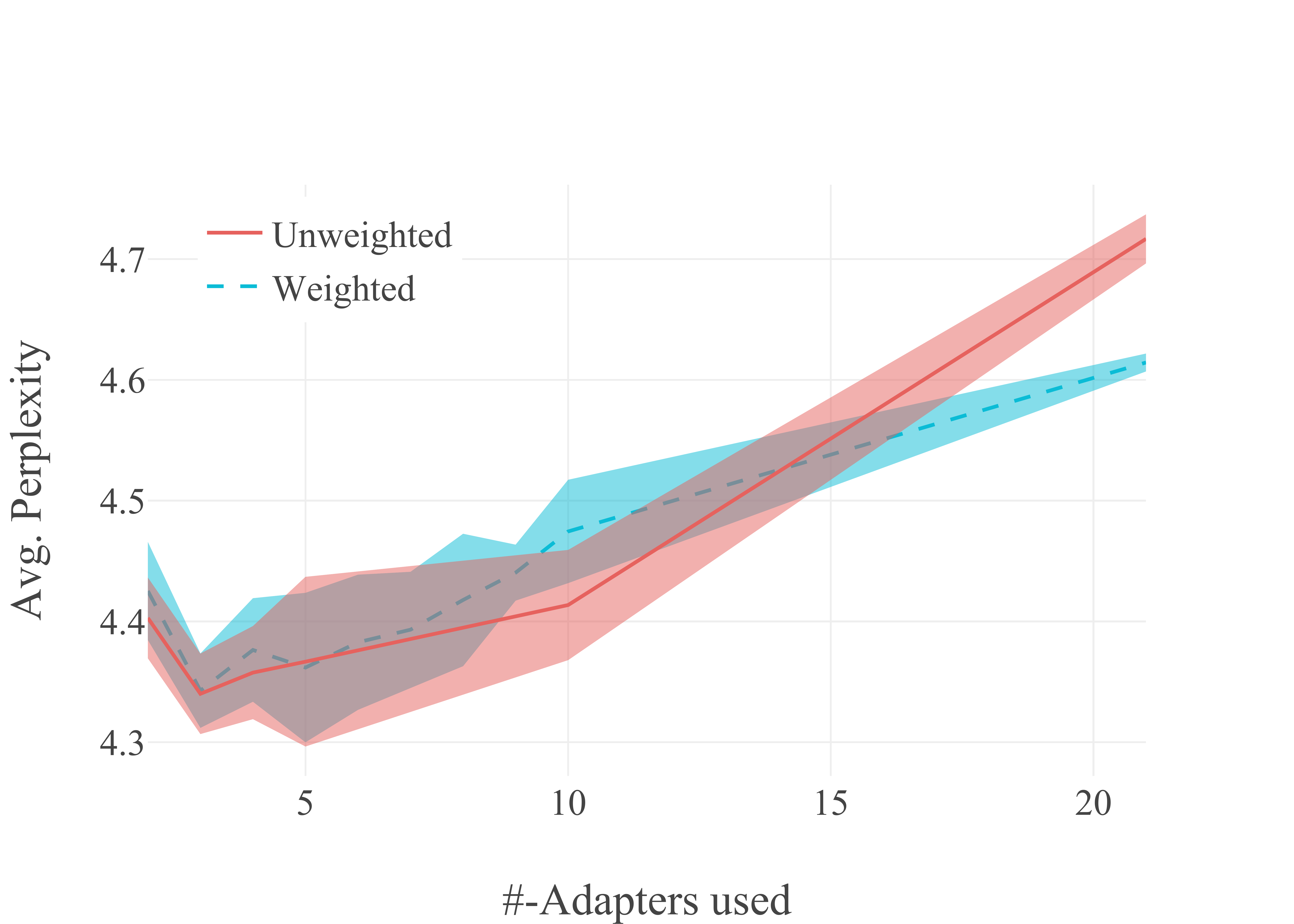}
         \caption{\textsc{tf--idf} - ensemble}
     \end{subfigure}
     \hfill
     \begin{subfigure}[b]{0.3\textwidth}
         \centering
         \includegraphics[width=\textwidth]{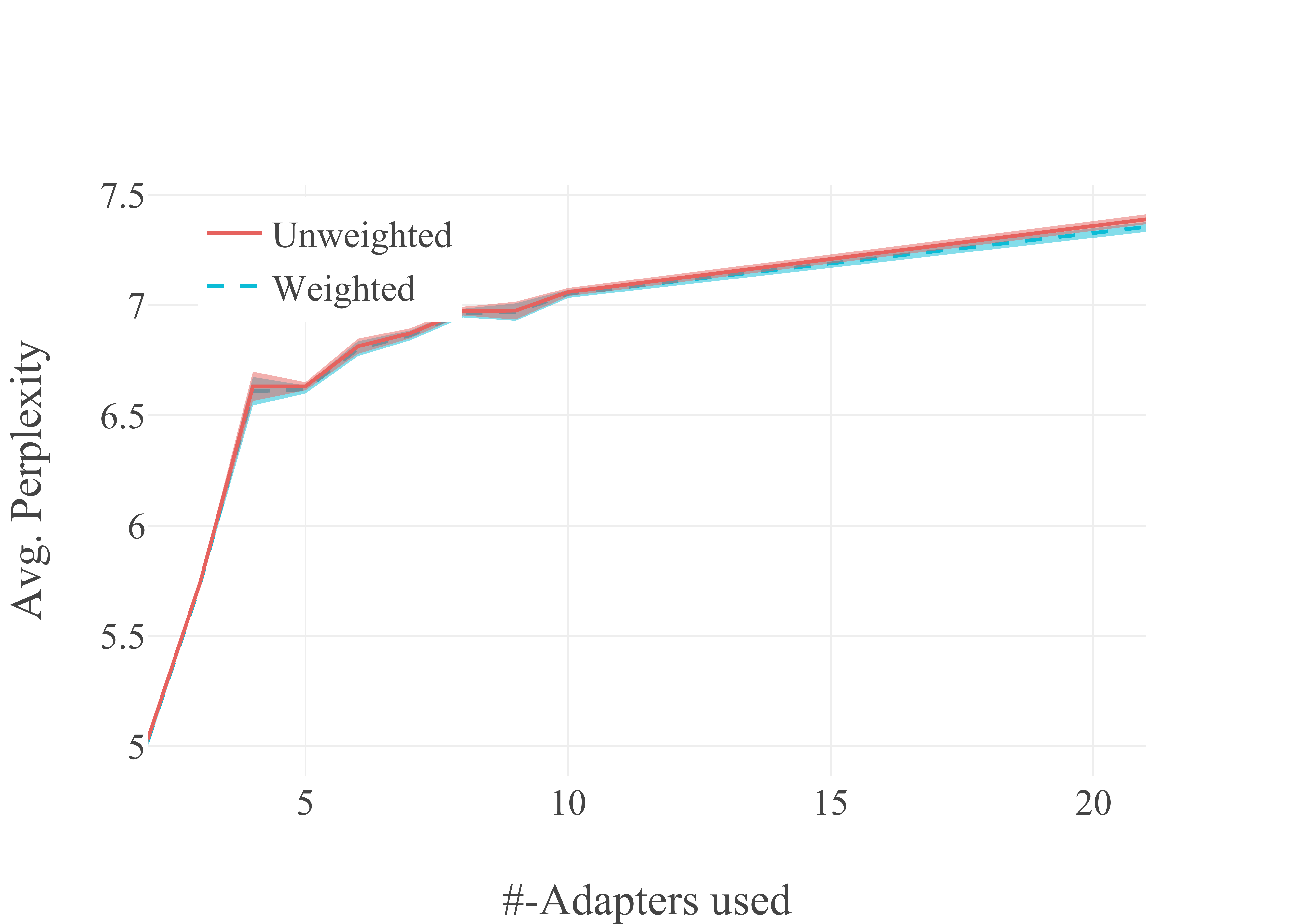}
         \caption{\textsc{entropy} - average}
     \end{subfigure}
     \hfill
     \begin{subfigure}[b]{0.3\textwidth}
         \centering
         \includegraphics[width=\textwidth]{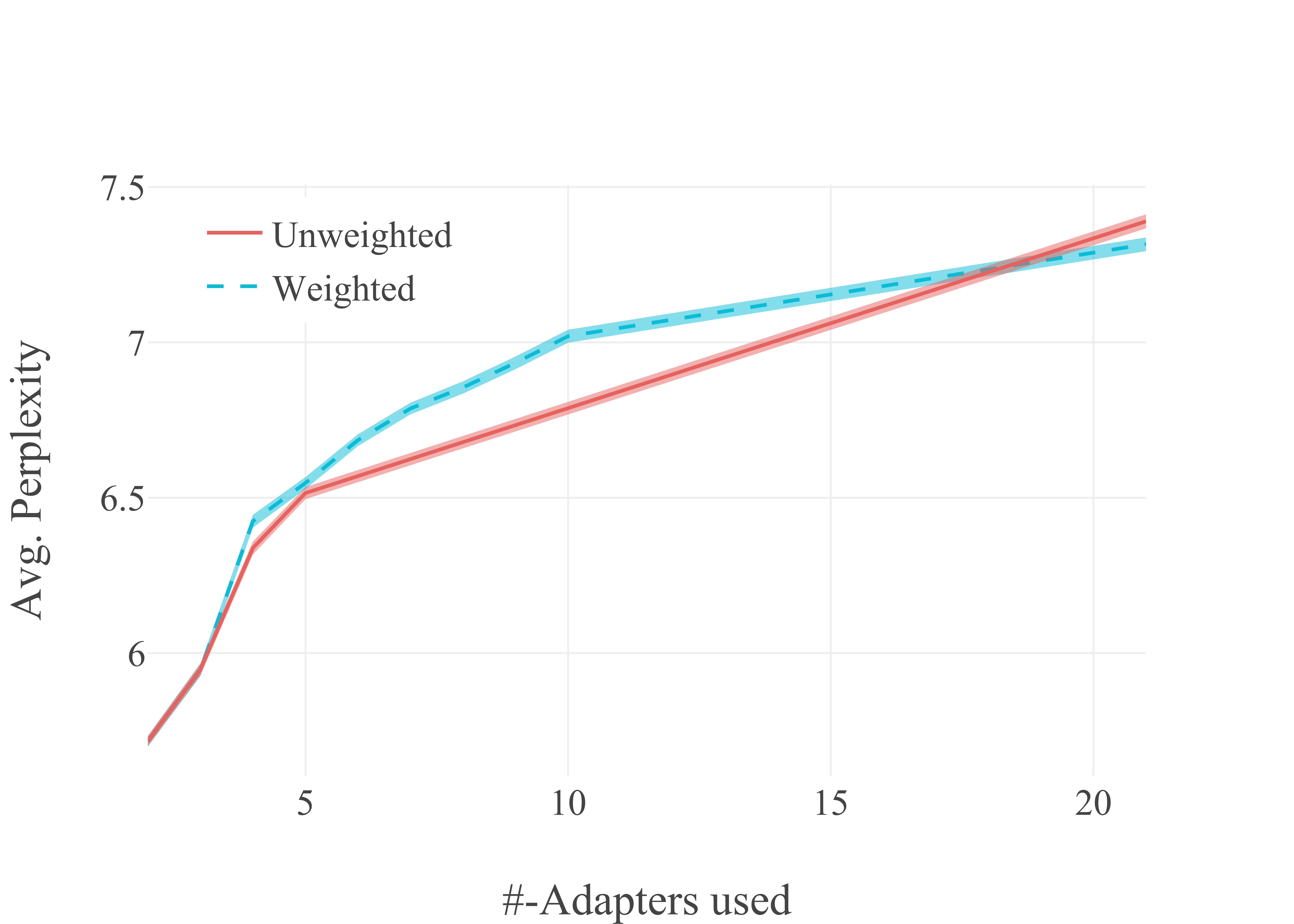}
         \caption{\textsc{prior} - average}
     \end{subfigure}
     \hfill
     \begin{subfigure}[b]{0.3\textwidth}
         \centering
         \includegraphics[width=\textwidth]{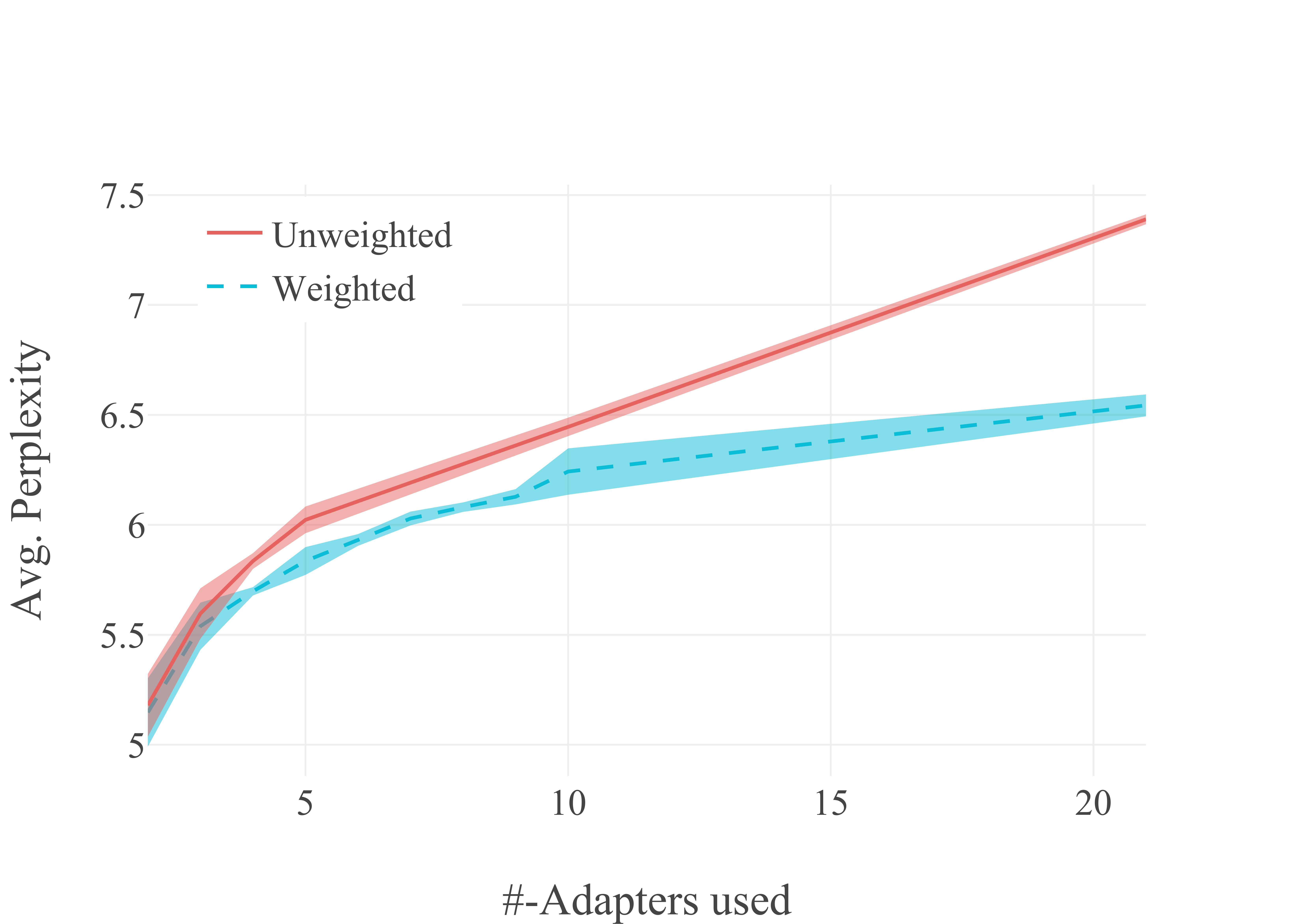}
         \caption{\textsc{SentSim} - average}
     \end{subfigure}
     \hfill
     \begin{subfigure}[b]{0.3\textwidth}
         \centering
         \includegraphics[width=\textwidth]{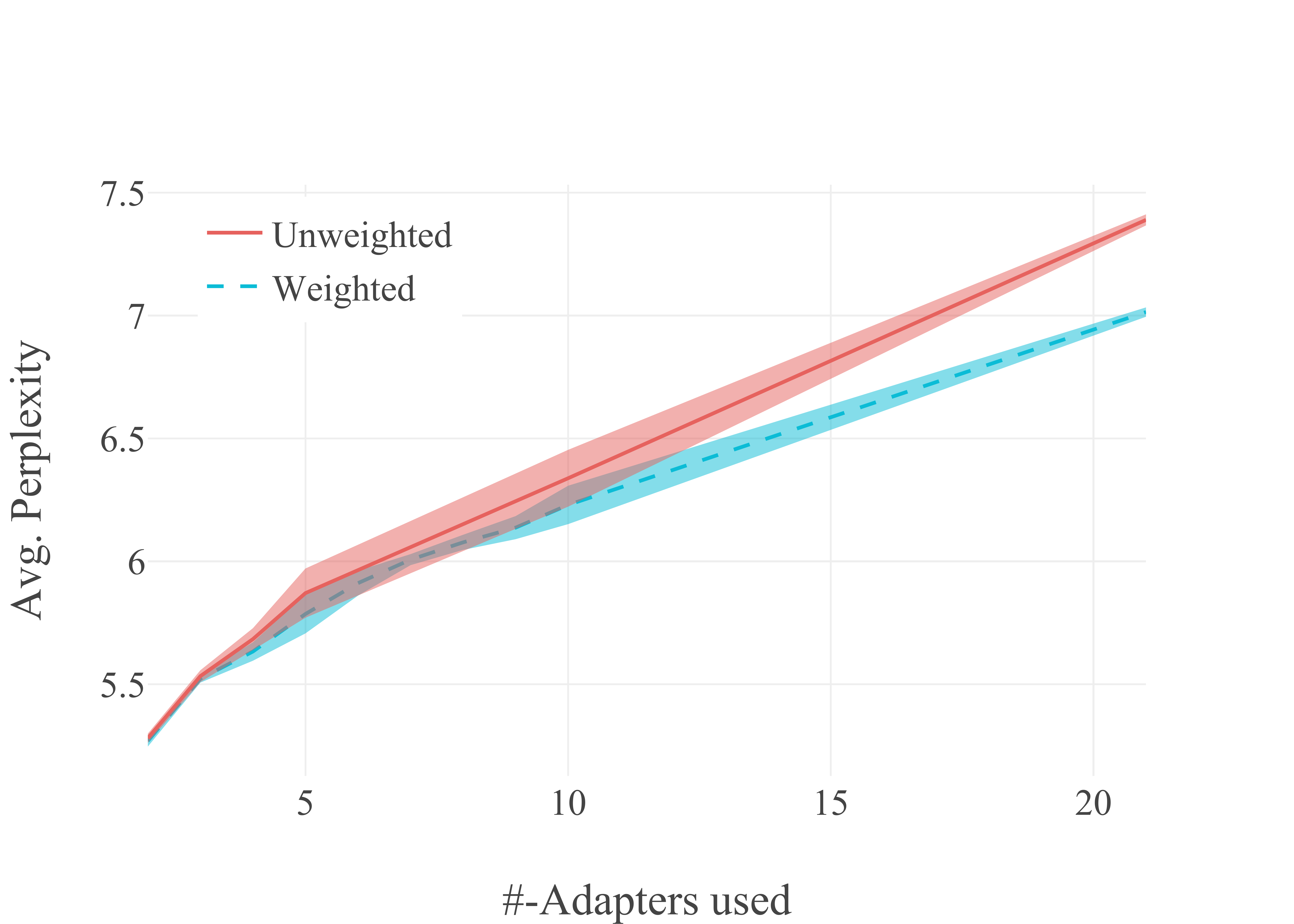}
         \caption{\textsc{tf--idf} - average}
     \end{subfigure}
     \hfill
      \caption[]{Comparison between weighting the selected adapters based on their similarity (blue) and assigning them uniform weights (red). We show the mean perplexity results averaged over all evaluation datasets and across four runs for {\small\texttt{deberta-base}} when using different pairings of scoring and combination strategies of our framework.}
\end{figure*}

\clearpage

\begin{figure*}[h]
     \centering
     \begin{subfigure}[b]{0.3\textwidth}
         \centering
         \includegraphics[width=\textwidth]{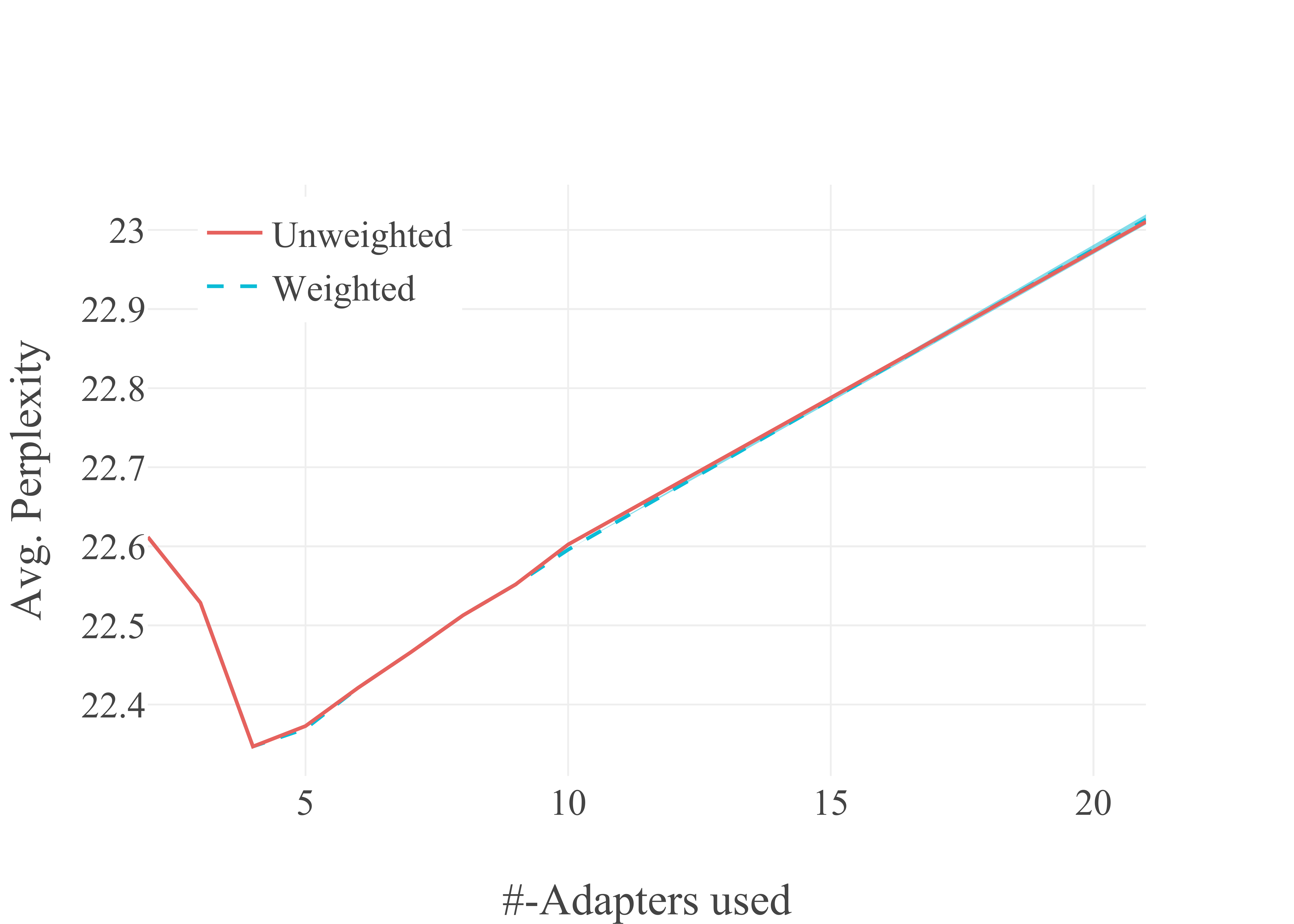}
         \caption{\textsc{entropy} - ensemble}
     \end{subfigure}
     \hfill
     \begin{subfigure}[b]{0.3\textwidth}
         \centering
         \includegraphics[width=\textwidth]{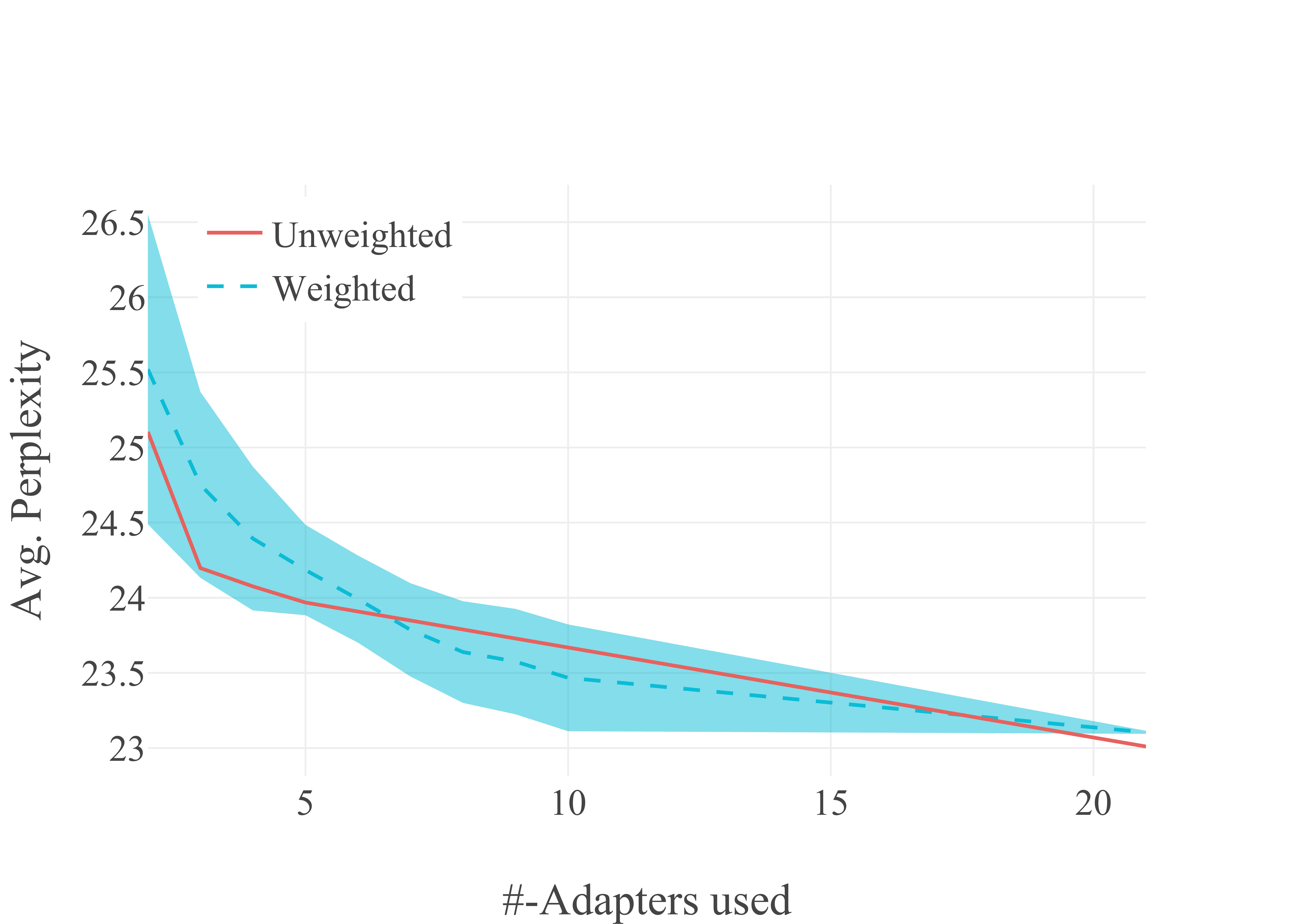}
         \caption{\textsc{prior} - ensemble}
     \end{subfigure}
     \hfill
     \begin{subfigure}[b]{0.3\textwidth}
         \centering
         \includegraphics[width=\textwidth]{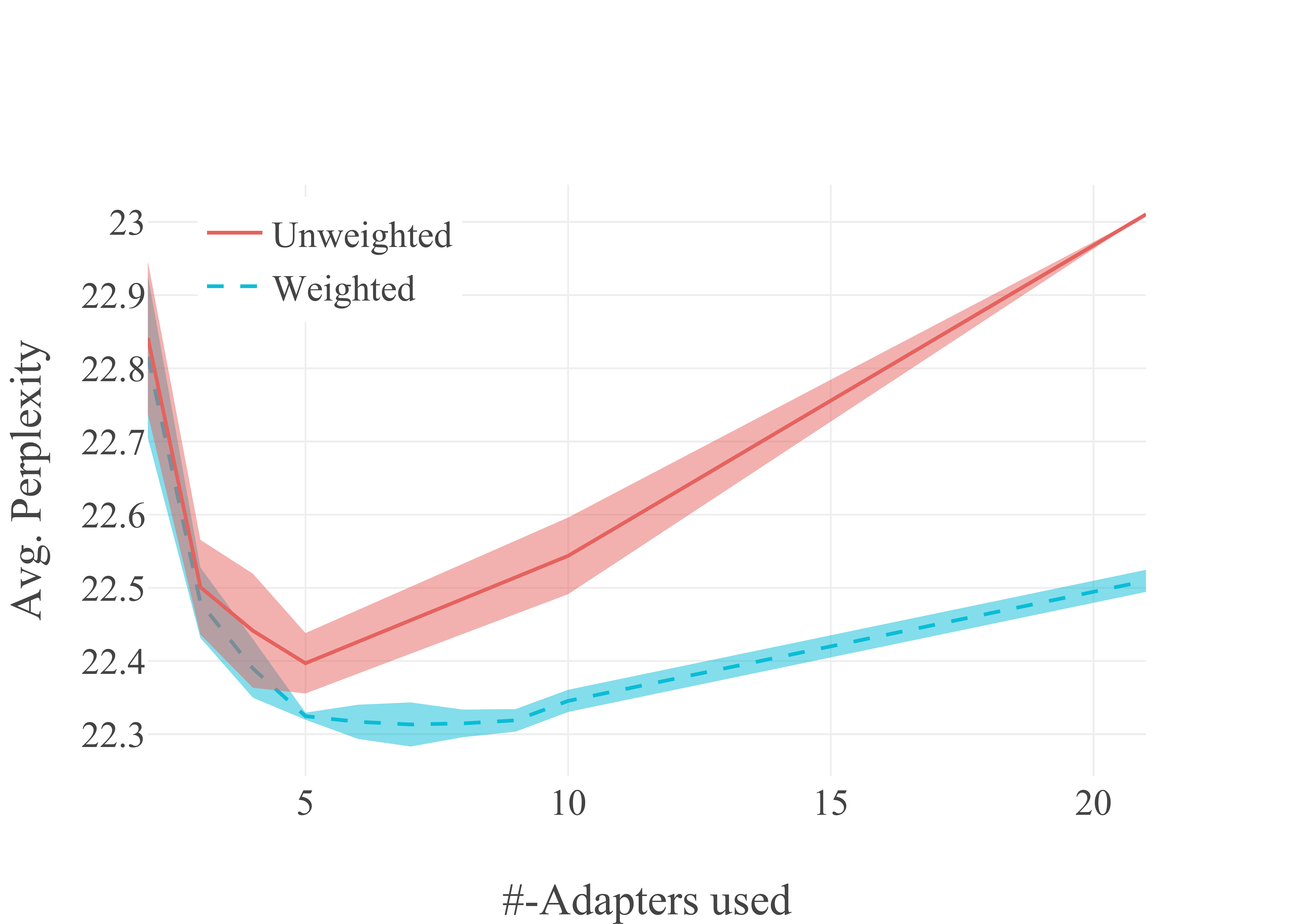}
         \caption{\textsc{SentSim} - ensemble}
     \end{subfigure}
     \hfill
     \begin{subfigure}[b]{0.3\textwidth}
         \centering
         \includegraphics[width=\textwidth]{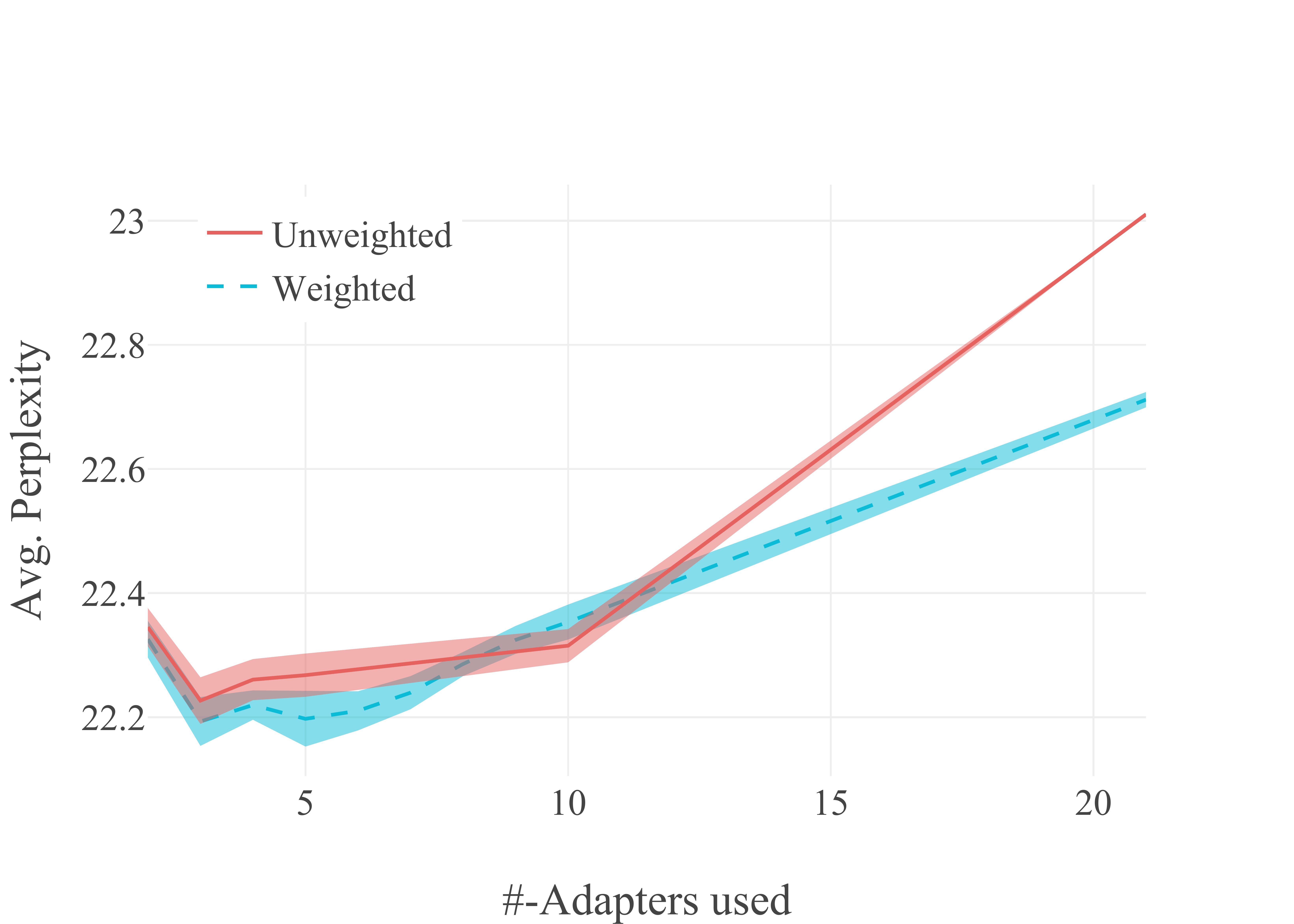}
         \caption{\textsc{tf--idf} - ensemble}
     \end{subfigure}
     \hfill
     \begin{subfigure}[b]{0.3\textwidth}
         \centering
         \includegraphics[width=\textwidth]{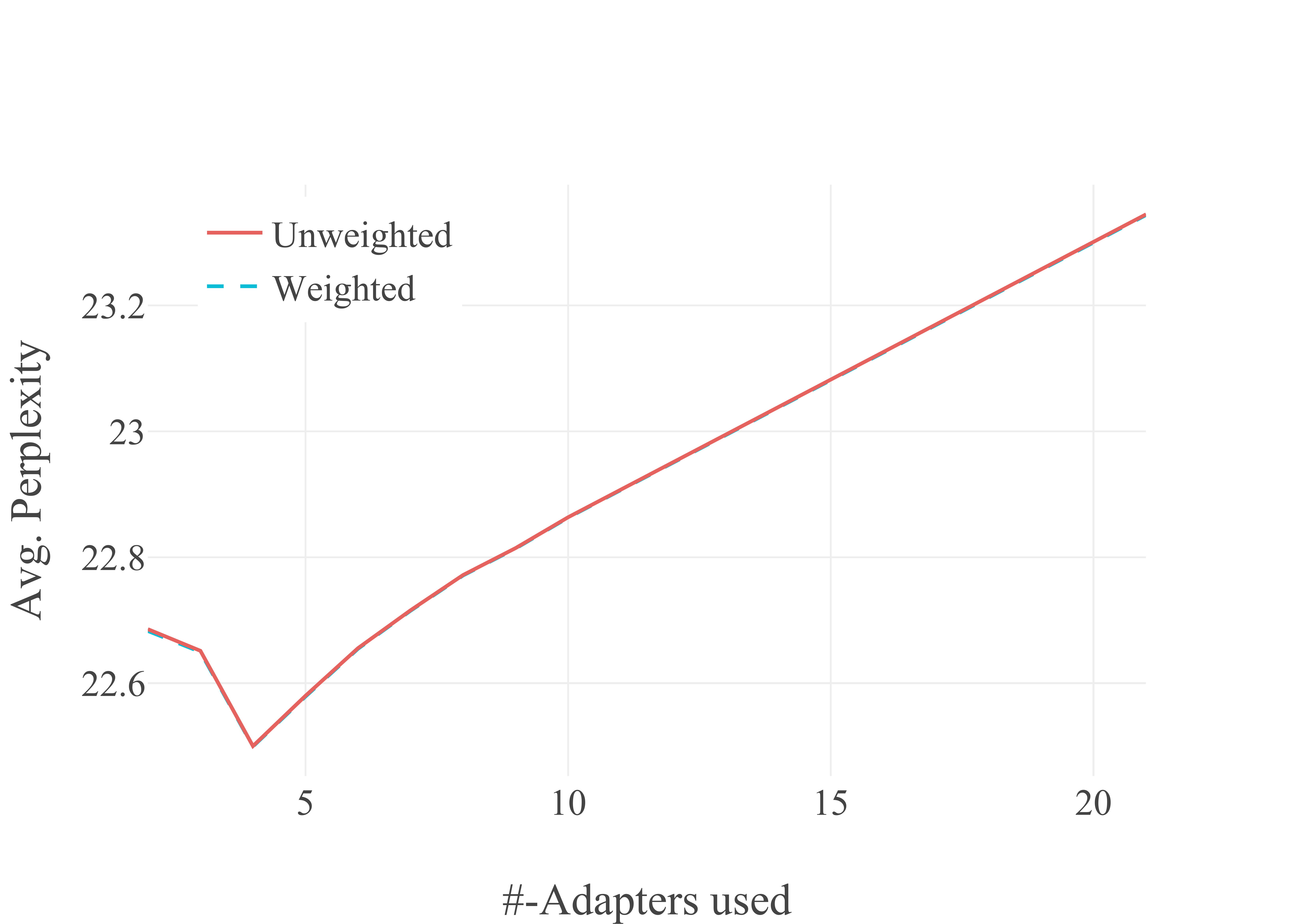}
         \caption{\textsc{entropy} - average}
     \end{subfigure}
     \hfill
     \begin{subfigure}[b]{0.3\textwidth}
         \centering
         \includegraphics[width=\textwidth]{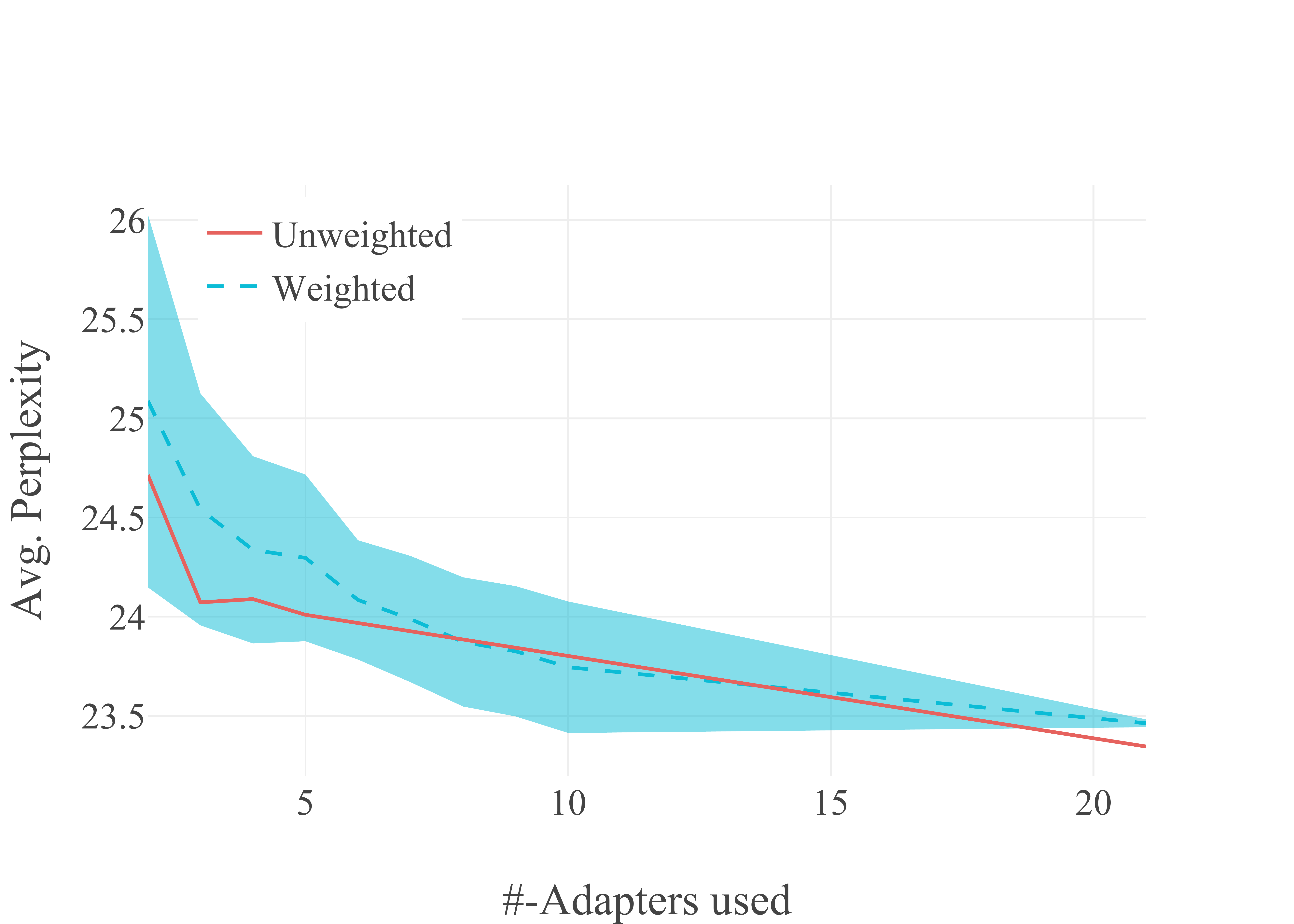}
         \caption{\textsc{prior} - average}
     \end{subfigure}
     \hfill
     \begin{subfigure}[b]{0.3\textwidth}
         \centering
         \includegraphics[width=\textwidth]{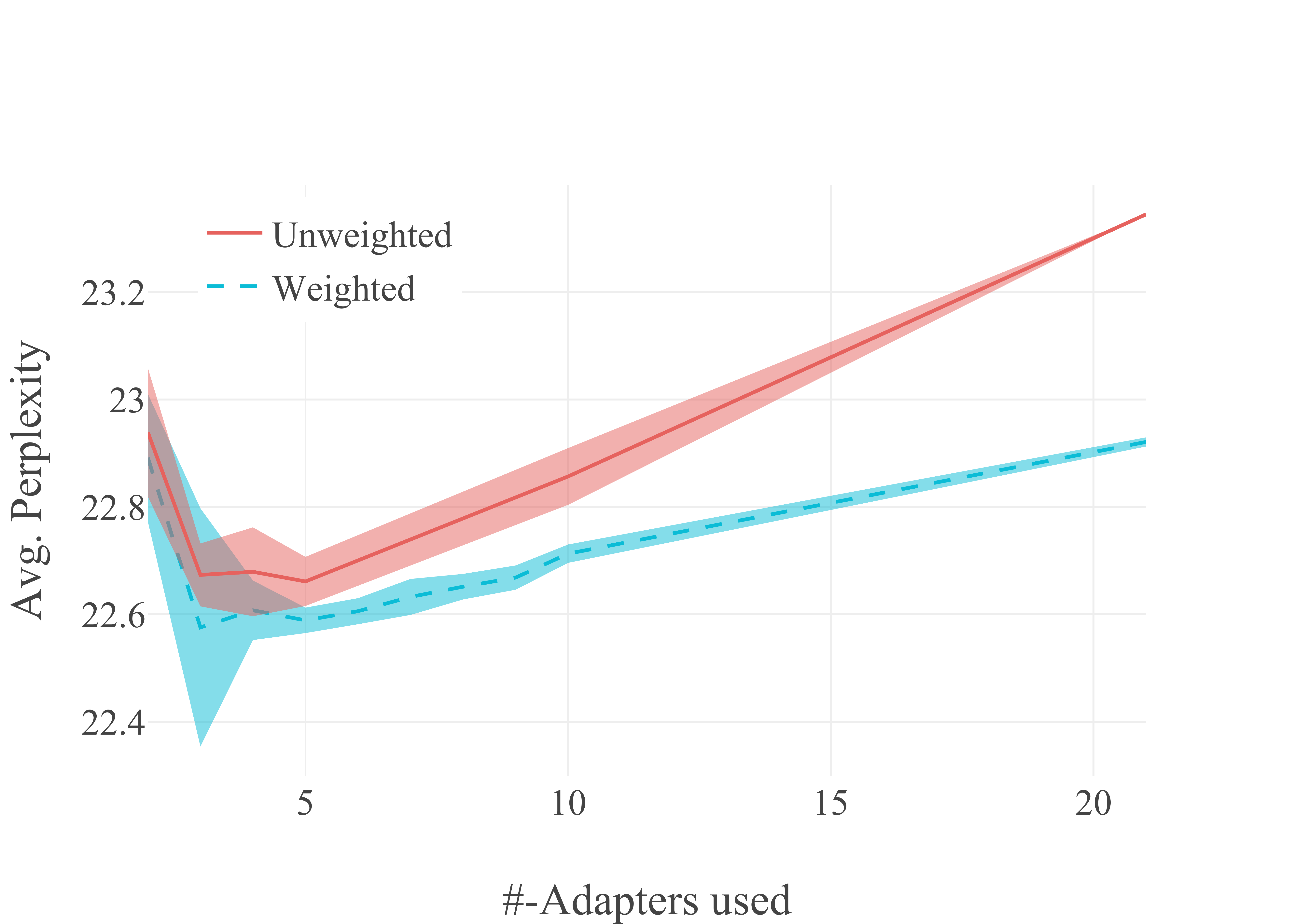}
         \caption{\textsc{SentSim} - average}
     \end{subfigure}
     \hfill
     \begin{subfigure}[b]{0.3\textwidth}
         \centering
         \includegraphics[width=\textwidth]{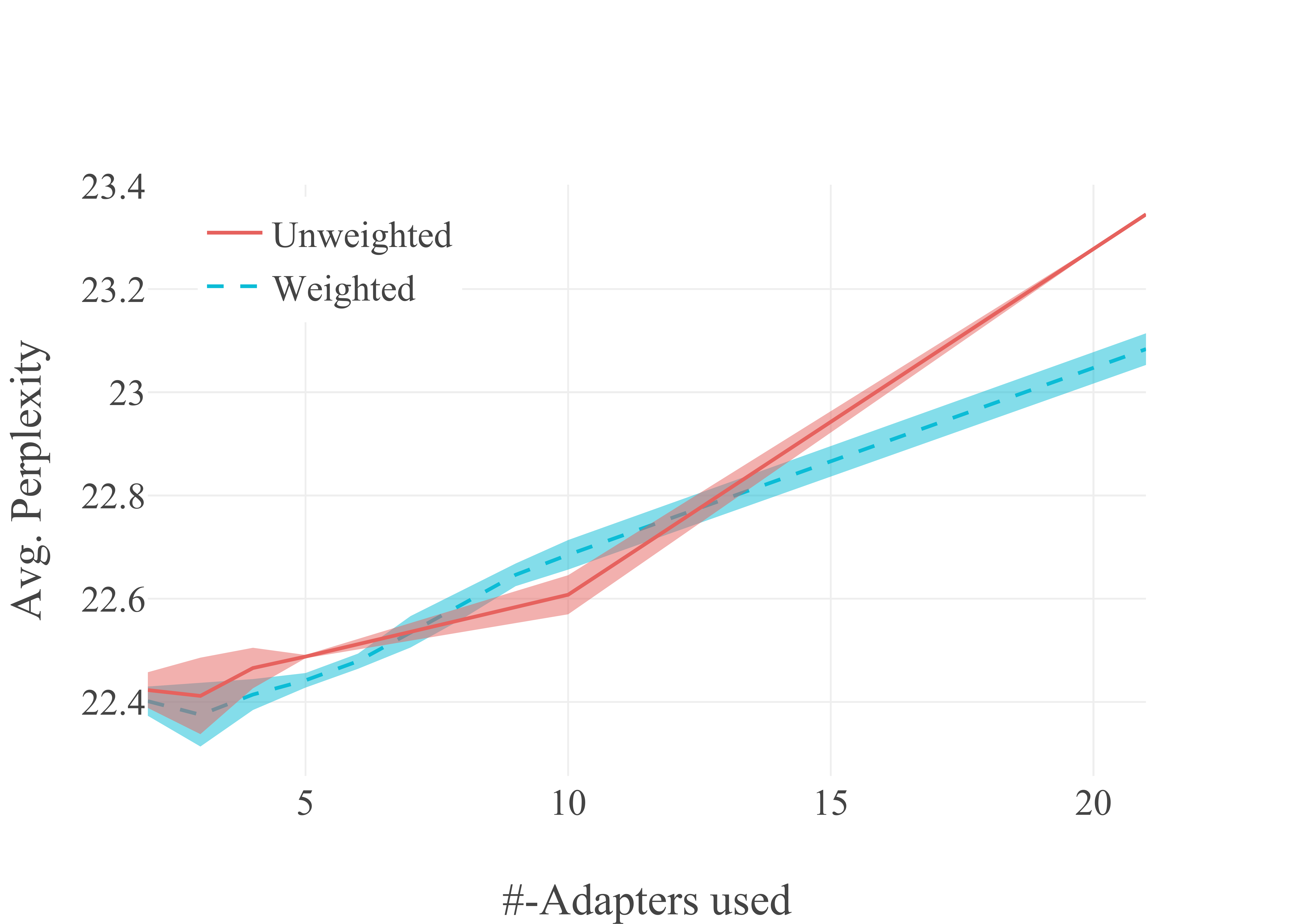}
         \caption{\textsc{tf--idf} - average}
     \end{subfigure}
     \hfill
      \caption[]{Comparison between weighting the selected adapters based on their similarity (blue) and assigning them uniform weights (red). We show the mean perplexity results averaged over all evaluation datasets and across four runs for {\small\texttt{gpt2-base}} when using different pairings of scoring and combination strategies of our framework.}
\end{figure*}

\clearpage

\twocolumn

\section{Efficiency of DeBERTa}\label{sec:efficiency_calc_app}
\begin{figure}[h]
    \centering
    \includegraphics[width=\linewidth]{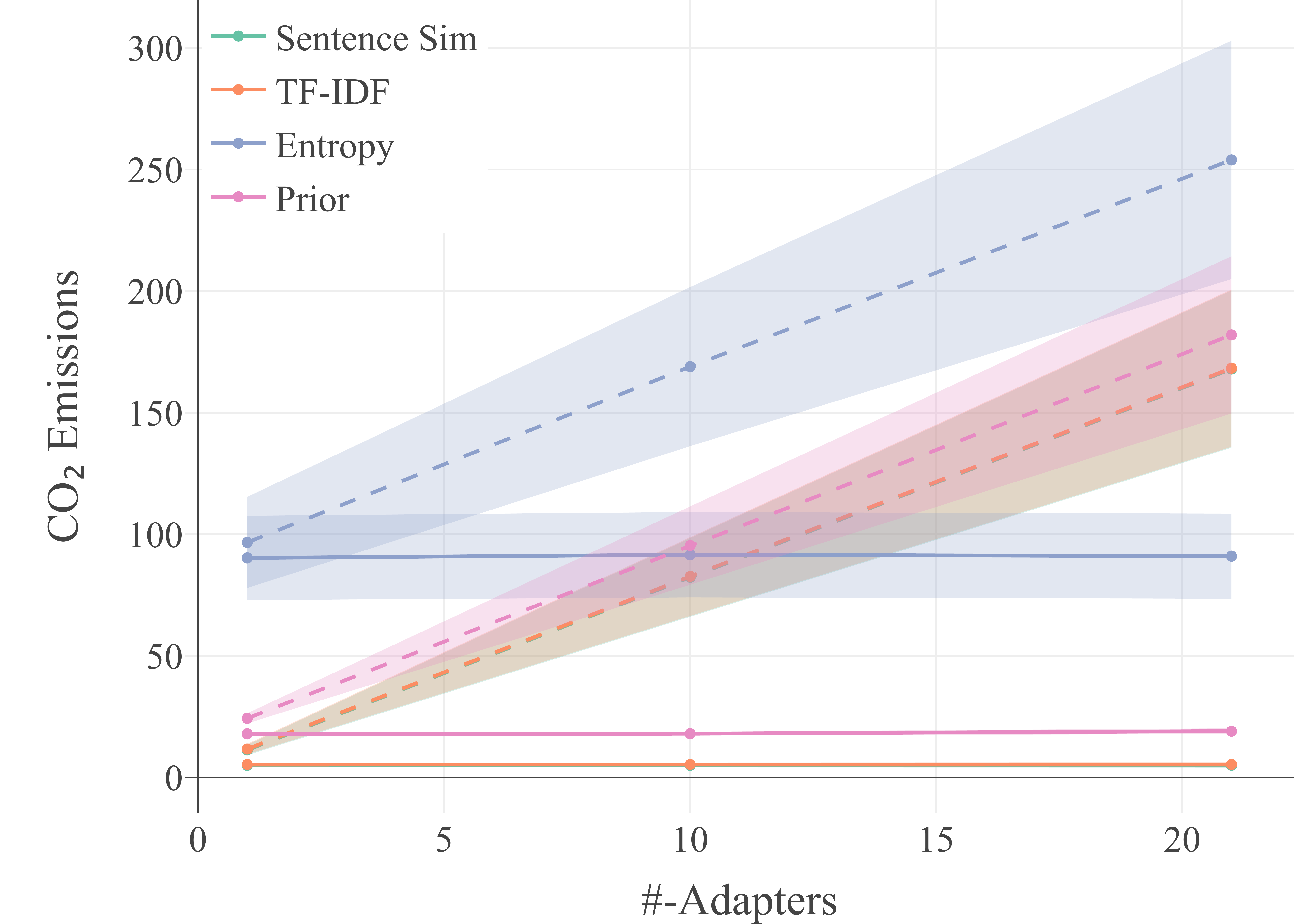}
    \caption{Comparison between the different selection and composition strategies with regards to their efficiency. We present the average CO$_2$Emissions for experiments where we conducted Parameter Averaging (solid lines) and Ensembling (dashed lines) over different numbers of top-$k$ adapters. We show the results for {\small\texttt{deberta-base}} when using each of our four scoring strategies (\textsc{SentSim}, \textsc{tf--idf}, \textsc{entropy}, \textsc{prior}) averaged across four runs.}
    \label{fig:carbon_emissions_deberta}
\end{figure}

We present the results of the efficiency calculations for {\small\texttt{deberta-base}} in Figure \ref{fig:carbon_emissions_deberta}. As expected, the plot shows the same pattern as for {\small\texttt{gpt2-base}}, with a linear increase in CO$_2$Emissions for a higher number of $k$.

\begin{table*}[t]
    \centering
    \small
    \begin{tabular}{|c|c|c|c|c|c|c|c|}
Threshold & \textsc{SentSim} - average &\textsc{tf--idf} - average & average & \textsc{SentSim} - ensemble & \textsc{tf--idf} - ensemble & ensemble & Total\\
\hline
0.001   & 0.64  &  0.84  &  0.74  &  0.55  &  0.73  &  0.64 &   0.69  \\ 
0.002   &  0.64  &  0.84  &  0.74  &  0.55  &  0.73  &  0.64 &   0.69  \\ 
0.003   &  0.67  &  \textbf{0.88}  &  0.77  &  0.57  &  0.79  &  0.68 &   0.73  \\ 
0.004   &  0.78  &  \textbf{0.88}  &  \textbf{0.83}  &  0.70  &  \textbf{0.80}  &  \textbf{0.75} &   \textbf{0.79}  \\ 
0.005   &  \textbf{0.79}  &  0.82  &  0.80  &  \textbf{0.73}  &  0.77  &  \textbf{0.75} &   0.78  \\ 
0.006   &  0.74  &  0.79  &  0.77  &  0.69  &  0.78  &  0.74 &   0.75  \\ 
0.007   &  0.74  &  0.74  &  0.74  &  0.69  &  0.73  &  0.71 &   0.73  \\ 
0.008   &  0.73  &  0.65  &  0.69  &  0.69  &  0.68  &  0.69 &   0.69  \\ 
0.009   &  0.73  &  0.42  &  0.57  &  0.69  &  0.47  &  0.58 &   0.58  \\ 
0.01   &  0.75  &  0.42  &  0.58  &  0.72  &  0.47  &  0.60 &   0.59 \\

    \end{tabular}
    \caption{Results for threshold tuning for an automatic selection of the best value for $k$. We show the percentage of how close we can get to the optimal value of $k$ with the respective threshold. We present the average of this percentage over each scoring strategy (\textsc{tf--idf} and \textsc{SentSim}) paired with each combination strategy, each combination strategy alone, and overall (Total).}
    \label{tab:hyperparameter_diff_thres}
\end{table*}

\section{Threshold Tuning via Early Stopping}\label{sec:thres}

In this additional experiment, we tried to estimate the optimal number of adapters to select by applying an early stopping algorithm, whenever we see a sudden drop in adapter similarity.

\begin{figure}[h]
    \centering
    \includegraphics[width=\linewidth]{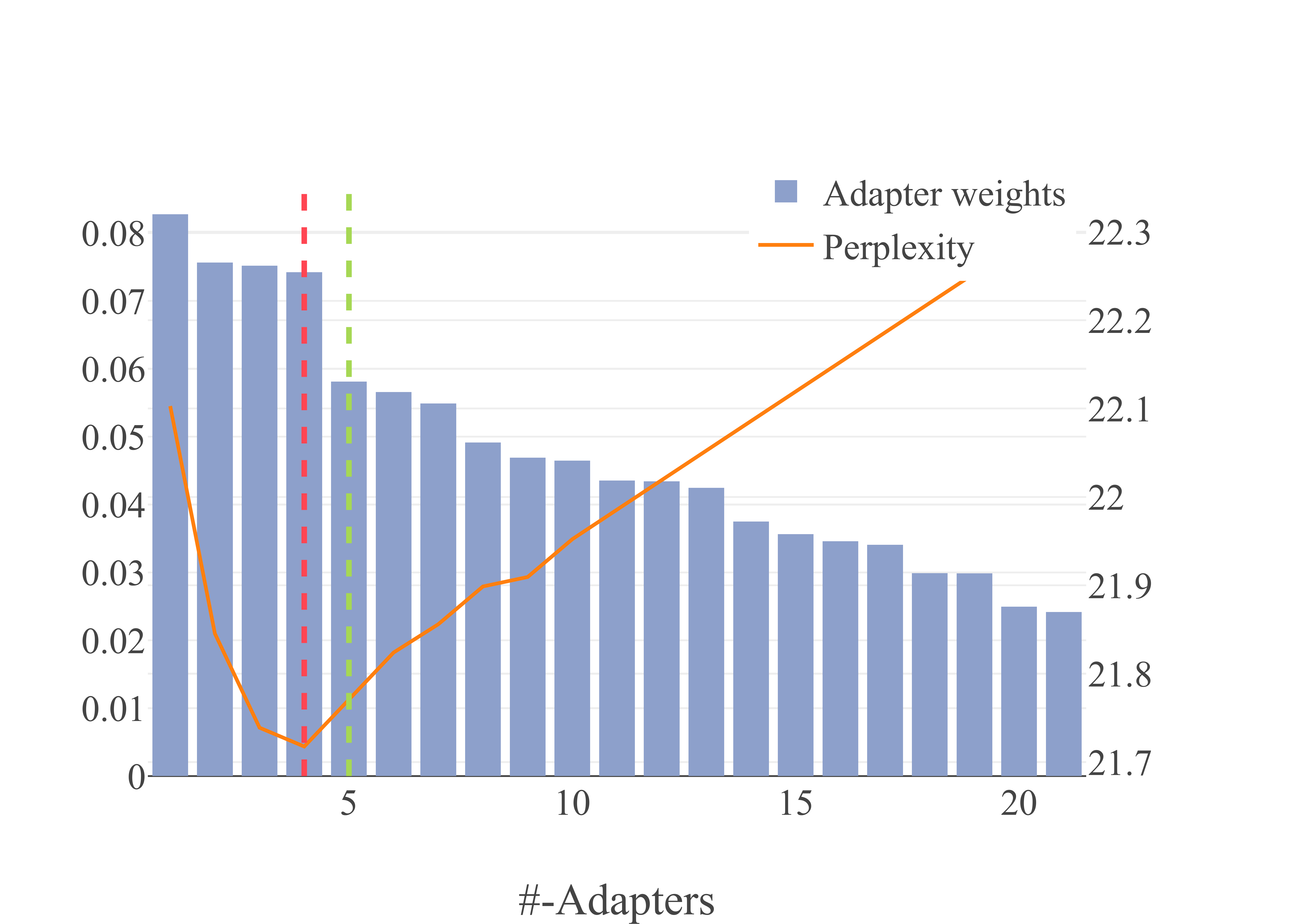}
    \caption{Visualization of the early stopping approach. The red vertical line marks the adapter combination leading to the result with the lowest perplexity. The vertical green line marks the number of adapters that would be chosen when applying the early stopping mechanism. The orange line shows the perplexity change when adding more adapters for this strategy. In this case, we show the results for {\small\texttt{gpt2-base}} on the techcrunch domain using \textsc{tf--idf} and ensemble the output.}
    \label{fig:early_stopping}
\end{figure}

For this experiment, we use the weighting strategies using \textsc{tf--idf} and \textsc{SentSim}, since these exhibited the largest variation in similarity weights. We then sort these weights from largest to smallest representing the adapter with the respective importance for the novel evaluation domain. We then iterate over the adapter weights and stop if the difference between the weights is larger than a certain threshold.  
We illustrate this procedure in Figure \ref{fig:early_stopping}. We run several experiments with different values set for the stopping threshold (see Table \ref{tab:hyperparameter_diff_thres}) and find that with a threshold of 0.004, we are able to obtain on average over all datasets and combination strategies 79\% of the optimal model performance.

\end{document}